\documentclass[acmtog,table]{acmart}
\acmSubmissionID{1149}
\usepackage{enumitem}
\usepackage{booktabs} 
\usepackage{xcolor}
\usepackage[most]{tcolorbox}
\usepackage{pifont}
\usepackage{subcaption}
\usepackage{ulem}
\newcommand{\cmark}{\ding{51}}
\newcommand{\xmark}{\ding{55}}

\definecolor{ex1figure_red}{RGB}{200,29,49}
\definecolor{ex1figure_blue}{RGB}{46,84,161}

\definecolor{ex2figure_red}{RGB}{203,41,60}

\definecolor{ex2figure_blue}{RGB}{47,85,151}

\usepackage{array}
\usepackage{booktabs}
\usepackage{graphicx}
\usepackage[ruled]{algorithm2e} 
\usepackage{tabularx}
\usepackage{booktabs}

\usepackage{multirow} 

\SetAlFnt{\small}
\SetAlCapFnt{\small}
\SetAlCapNameFnt{\small}
\SetAlCapHSkip{0pt}

\acmJournal{TOG} 
\acmConference[]{}{}{}
\acmDOI{}
\acmISBN{}

\begin{document}
\begin{teaserfigure}
  \includegraphics[width=0.99\textwidth]{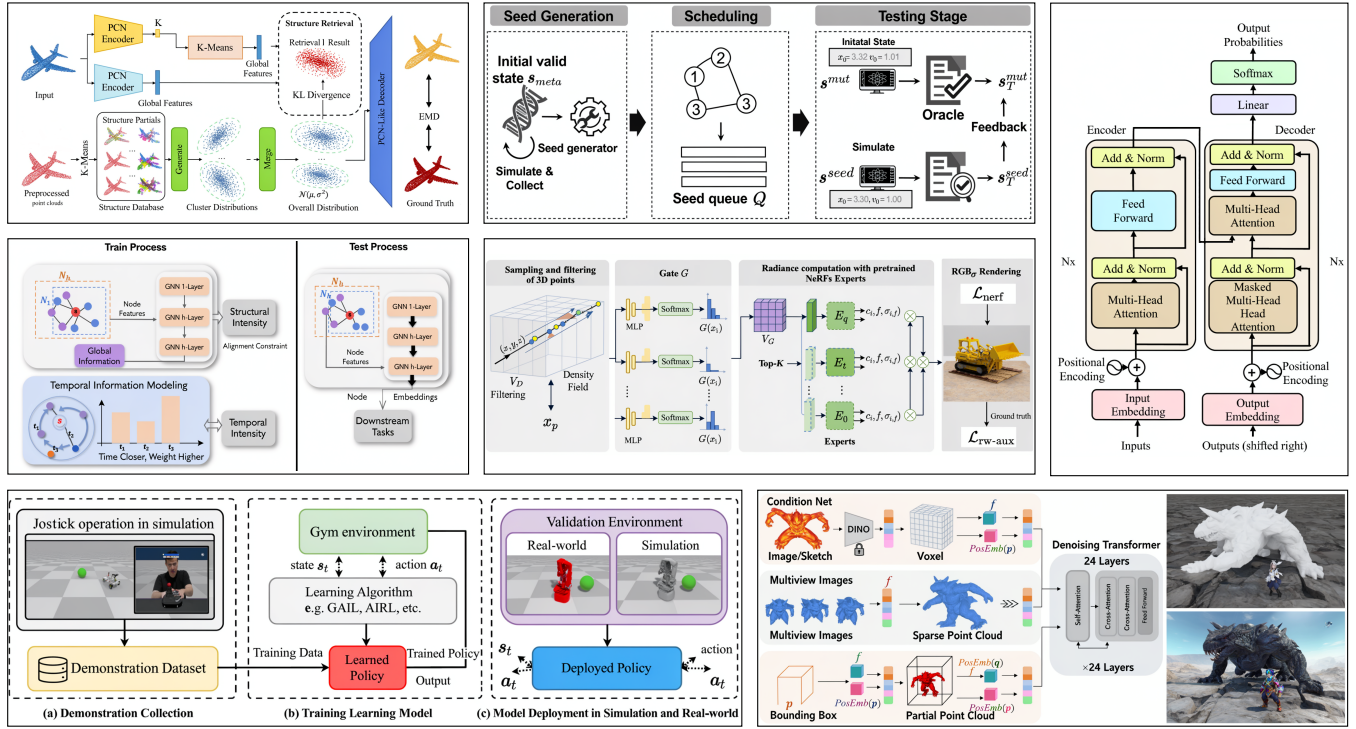}
  \vspace{-2mm}
  \caption{Diverse scientific methodology diagrams generated by SciForma. Our framework ensures strict \textit{structural fidelity}, enables precise synthesis of complex topologies, dense block layouts, and accurate textual annotations across various academic domains.}
  \Description{Teaser image}
  \label{fig:teaser}
\end{teaserfigure}

\title{SciForma: Structure-Faithful Generation of Scientific Diagrams}


\author{Yuxuan Luo$^*$}
\thanks{$^*$ This work was done when Yuxuan Luo interned at Microsoft Research Asia.}
\orcid{0009-0004-7363-356X}
\affiliation{
    \institution{Wangxuan Institute of Computer Technology, Peking University}
    \city{Beijing}
    \country{China}
}

\author{Peng Zhang}
\orcid{0000-0003-2246-9080}
\affiliation{
    \institution{State Key Lab of CAD \& CG, Zhejiang University}
    \city{Hangzhou}
    \country{China}
}

\author{Xinjie Zhang$\dagger$}
\orcid{0000-0002-3194-7518}
\thanks{$\dagger$ Corresponding authors.} 
\affiliation{
    \institution{Microsoft Research Asia}
    \country{\relax}
}
\email{xinjiezhang@microsoft.com}

\author{Xun Guo}
\orcid{0009-0005-4627-401X}
\affiliation{
    \institution{Microsoft Research Asia}
    \country{\relax}
}

\author{Zhouhui Lian$\dagger$} 
\orcid{0000-0002-2683-7170}
\affiliation{
    \institution{Wangxuan Institute of Computer Technology, Peking University}
    \city{Beijing}
    \country{China}
}
\email{lianzhouhui@pku.edu.cn}

\author{Yan Lu}
\orcid{0000-0001-5383-6424}
\affiliation{
    \institution{Microsoft Research Asia}
    \country{\relax}
}

\begin{abstract}
\textit{Structural fidelity} is essential to scientific methodology diagrams. To communicate research logic, these diagrams must faithfully render components, directional relations, and textual annotations. Since a single error, such as a reversed arrow or an unreadable equation, can invalidate the entire figure, \textit{structural fidelity} is inherently conjunctive: correctness on one axis cannot compensate for failure on another. Current open-source models fail to satisfy this criterion. Supervised fine-tuning (SFT) learns plausible layouts but cannot reliably ensure structural correctness, while scalar reward-based post-training obscures which structural dimension has failed. To address this, we introduce \textbf{SciForma}, a framework for the structure-faithful generation of scientific methodology diagrams. Specifically, SciForma decomposes diagram quality into three structural axes: Component, Arrow, and Text, guided by a \textbf{structural inventory}. Built on this foundation, we curate \textbf{SciFormaData-700K} for structured training and \textbf{SciFormaBench-2K} for logic-verified evaluation. To close the gap left by SFT, we develop \textbf{Multi-Dimensional Conjunctive Preference Optimization (M-DPO)}, which enforces simultaneous correctness across all axes and adaptively routes gradients to the most deficient dimension in post-training. The same structural inventory also enables iterative editing at inference time to correct residual errors. This combination allows \textbf{SciForma-9B} to exceed all open-source baselines and GPT-Image-1.5 on both SciFormaBench-2K and AIBench, bringing open scientific-diagram generation close to proprietary-level \textit{structural fidelity}. Our code and data will be available at: \url{https://github.com/microsoft/SciForma}.

\end{abstract}
\authorsaddresses{Corresponding authors: Xinjie Zhang (xinjiezhang@microsoft.com), Microsoft Research Asia; Zhouhui Lian (lianzhouhui@pku.edu.cn), Wangxuan Institute of Computer Technology, Peking University.}
%
%

%
%


\maketitle
\section{Introduction}
\label{sec:intro}
Methodology diagrams, such as process pipelines and architecture overviews, serve as blueprints in scientific papers, making research logic visible and verifiable. Unlike natural images that prioritize visual plausibility, methodology diagrams demand \textit{structural fidelity}: every component, directional relation, and textual annotation must be legible. A reversed arrow or an unreadable equation does not merely degrade visual quality; it can invalidate the meaning of the diagram. \textit{Structural fidelity} is therefore non-compensatory: no strength on one axis can offset a failure on another.


Current open-source models trail proprietary systems like~\cite{NanoBananaPro, GPTImage2} on \textit{structural fidelity}. This gap reflects a limitation of current training objectives. Supervised fine-tuning (SFT) learns plausible distributions through flow-matching loss, but offers no guarantee of complete structural correctness. Meanwhile, existing post-training methods~\cite{GRPO, GDRO, DPO} collapse orthogonal structural axes into a single scalar signal, obscuring which dimension has failed. At root, current approaches lack mechanisms to independently verify and enforce correctness on each structural axis.


To address this mismatch, we introduce \textbf{SciForma}, a framework for structure-faithful generation of scientific methodology diagrams. Following established process-modeling standards~\cite{UML25,BPMN2}, our key insight is that diagram quality is not a single holistic property, but decomposes into three independently verifiable structural axes: semantic modules (\textbf{C}omponents), directional relations (\textbf{A}rrows), and textual annotations (\textbf{T}ext). We formalize this evaluation through a \textbf{structural inventory} that extracts ground-truth checklists from reference diagrams for per-axis verification. This decomposition recasts the task from continuous visual matching to explicit structural verification, so failures in one axis can no longer be obscured by strengths in others.


To sum up, major contributions of this paper include:

\begin{itemize}[leftmargin=*, topsep=1pt, itemsep=2pt, parsep=0pt]
\item \textbf{Data Foundation.} We curate \textbf{SciFormaData-700K}, a topology-captioned dataset for structured generation and editing, alongside \textbf{SciFormaBench-2K} for logic-verified evaluation.

\item \textbf{Multi-Dimensional Conjunctive Preference Optimization (M-DPO).} To close the gap left by SFT, we develop M-DPO. It contrasts a global winner against axis-specific losers via a multi-way Bradley–Terry objective, enforcing conjunctive correctness while adaptively routing gradients toward the most deficient axis.

\item \textbf{SciForma.}  With this data we fine-tune \textbf{SciForma-Base}, which is then post-trained into \textbf{SciForma-9B}. At inference time, the same structural inventory enables an iterative refinement to identify and correct residual errors through verification-gated inpainting.
\end{itemize}

On \textbf{SciFormaBench-2K}, SciForma-9B exceeds all open-source baselines and GPT-Image-1.5; M-DPO delivers its largest gains on the Arrow axis and the Hard tier. On the VQA-centric \textbf{AIBench}, it exceeds human-drawn originals with the largest margin on Topology. Integrated as a Visualizer in \textbf{PaperBanana}, it generalizes to agentic-generated prompts. Our ablations confirm the efficacy of M-DPO: (i) it breaks through the SFT stagnation; (ii) it avoids the reward collapse inherent in scalar-based methods (e.g., GDRO/GRPO); and (iii) it demonstrates that the coupling of decomposed pairs with a conjunctive objective is vital for consistent cross-axis gains.

\begin{figure*}
    \centering
    \includegraphics[width=0.995\linewidth]{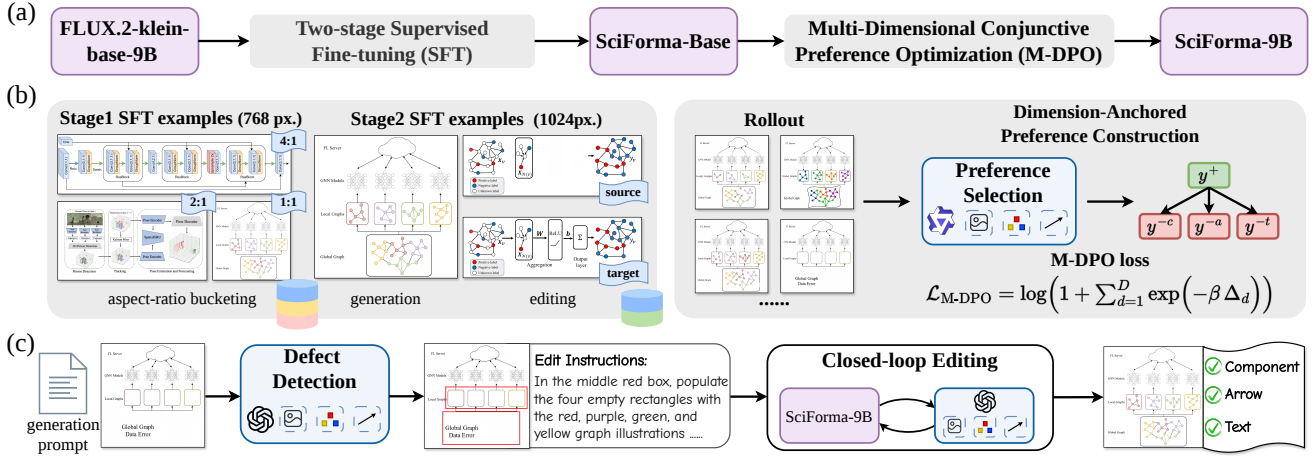}
    \Description{A three-part overview of the SciForma framework. Part (a) shows the training pipeline from Base to 9B. Part (b) illustrates the data formats for SFT and the M-DPO optimization logic. Part (c) details the iterative refinement loop involving defect detection and localized editing.}
    \vspace{-4mm}
    \caption{Overview of SciForma. \textbf{(a) Pipeline:} Two-stage SFT yields SciForma-Base, followed by M-DPO to produce SciForma-9B. \textbf{(b) Training Details:} \textit{Left:} SFT data formats for joint generation and editing. \textit{Right:} M-DPO optimizes a shared winner against dimension-anchored losers to model conjunctive preference. \textbf{(c) Iterative Refinement:} A closed-loop inference pipeline detects structural defects and applies localized editing to improve \textit{structural fidelity}.}
    \vspace{-2mm}
    \label{fig:overview}
\end{figure*}

\section{Related Work}

\subsection{Scientific Diagram Generation}

\noindent\textbf{Code-based generation} synthesizes diagrams via structured languages like TikZ~\cite{Automatikz,Detikzify,Tikzero,TikZilla}, Mermaid/SVG~\cite{SciSketch,SVGen,DiagramEval}, or domain-specific markup~\cite{SciDoc2Diagrammer}. While these approaches offer precise control, their expressiveness is inherently constrained, resulting in simple or template-driven visualizations. 



\noindent\textbf{Image-based generation} for scientific diagrams remains challenging. Prior work~\cite{Figgen}, open-source models~\cite{FLUX1dev,QwenImage}, and unified generators~\cite{zhao2025unified,wang2025ovis,callireader} struggle with \textit{structural fidelity} and topological consistency~\cite{SciFlow-Bench,AIBench,Mmmg}. Closed-source systems like Nano Banana Pro~\cite{NanoBananaPro} and GPT-Image-2~\cite{GPTImage2} demonstrate superior quality, but their training recipes remain unavailable.

\noindent\textbf{Agentic frameworks}~\cite{Diagrammergpt,SciFig,Paper2SysArch,DiagrammerAgent,Autofigure,AutofigureEdit,PaperBanana} employ retrieval, auditing, and self-reflection to improve on benchmarks~\cite{AIBench,SciFlow-Bench,ProImage-Bench,GENFIG1}. However, these gains are obtained at inference time through proprietary APIs, leaving the fundamental challenge of training such models to avoid fine-grained structural failures unaddressed. 
\vspace{-2mm}
\subsection{Post-training Alignment for Generative Models}

Post-training alignment for diffusion models falls into two families. \textbf{Offline methods} adapt DPO~\cite{DPO} to flow-matching via pairwise comparisons~\cite{DiffusionDPO,D3PO}, with extensions for ranking~\cite{MaPO,RankedPO} and multi-candidate ranking (GDRO~\cite{GDRO}). \textbf{Online methods} use policy-gradient~\cite{DDPO} or GRPO~\cite{GRPO}; subsequent work improves sampling efficiency~\cite{FlowGRPO,DanceGRPO} and training stability~\cite{DiffusionNFT,GRPOGuard,SmartGRPO}. Reward design has progressed from holistic scalar scores~\cite{ImageReward,HPSv2,PickScore} through rubric evaluation~\cite{VisionReward,RubricRewards} to multi-reward aggregation~\cite{CorrMultiReward,SampleRewardSoups}, yet all methods collapse per-axis signals into a single scalar objective.

\noindent\textbf{Alignment for structured synthesis} has focused on symbolic domains such as TikZ~\cite{TikZilla,DualSCRL}, SVG~\cite{RLRF}, text layout~\cite{TextDiffuserRL}, slide aesthetics~\cite{AeSlides}, and object-centric attributes~\cite{OSPO}, using domain-specific rewards for physics~\cite{PhysCorr} and 3D~\cite{DSO}. These methods optimize single-axis or scalar objectives and do not generalize to multi-axis structural verification.

\noindent\textbf{Multi-dimensional preference optimization} has recently appeared for natural image generation, including \textbf{CaPO}~\cite{lee2025calibrated} and \textbf{MCDPO}~\cite{jang2025multi}, along with others~\cite{li2025drpo, BalancedDPO,CPO}. These methods address perceptual axes (e.g., aesthetics) that co-vary smoothly: gains on one axis seldom degrade another, so scalar aggregation suffices. Neighbor GRPO~\cite{he2025neighbor} couples multiple candidates but still collapses feedback into a single scalar. However, none of these methods has been applied to independently verifiable structural axes under conjunctive constraints, where a single-axis failure invalidates the entire output. M-DPO addresses this limitation by enforcing conjunctive per-axis correctness.

\section{Structural Inventory}
\label{sec:schema}

To build a rigorous evaluation framework, we must answer two questions: what structural properties of a diagram are verifiable from a prompt and a reference image, and how can we measure them consistently at the element level? 
\subsection{Universal Structural Primitives}
\label{sec::why}

Scientific diagrams share a structural logic codified in established modeling standards. In UML activity diagrams~\cite{UML25} and BPMN process models~\cite{BPMN2}, communicative meaning is carried by typed entities, directed relations, and textual qualifiers.

Adopting this shared grammar, we formalize scientific methodology diagrams with three independently verifiable primitives: \textbf{Component(C)} for entity identity and spatial layout, \textbf{Arrow(A)} for information topology, and \textbf{Text(T)} for textual annotations. Together, these three primitives form the \textbf{structural inventory}, a ground-truth checklist that any structurally faithful output must satisfy. 

\subsection{Automated Inventory Extraction and Evaluation}
\label{sec:inventory}

The structural inventory makes diagram quality explicitly verifiable. Given a prompt and its reference image, a VLM extracts the structural inventory as a JSON checklist: \textbf{C}omponent coordinates, \textbf{A}rrow source-target pairs, and literal \textbf{T}ext strings. This checklist defines exactly what a structurally faithful output must contain.


We evaluate each diagram across three structural axes by comparing the output against the corresponding checklist. During this process, any identified discrepancy is labeled as either \emph{critical} ($w_e{=}1.0$, missing or fundamentally wrong) or \emph{moderate} ($w_e{=}0.5$, present but defective). Per-axis score $s_d$ is computed as:
\begin{equation}
    s_d = \max\Bigl( 1 - \frac{\sum_e w_e}{n_d} \Bigr),
\end{equation}
where $n_d$ is the total number of expected elements on axis $d$. While $\bar{s}$ (the per-axis average)  provides a compact summary, conjunctive correctness is assessed through the per-axis scores, which remain primary for diagnosis, reward construction, and ablation analysis.






\begin{figure*}[t!]
    \centering
    \includegraphics[width=0.96\linewidth]{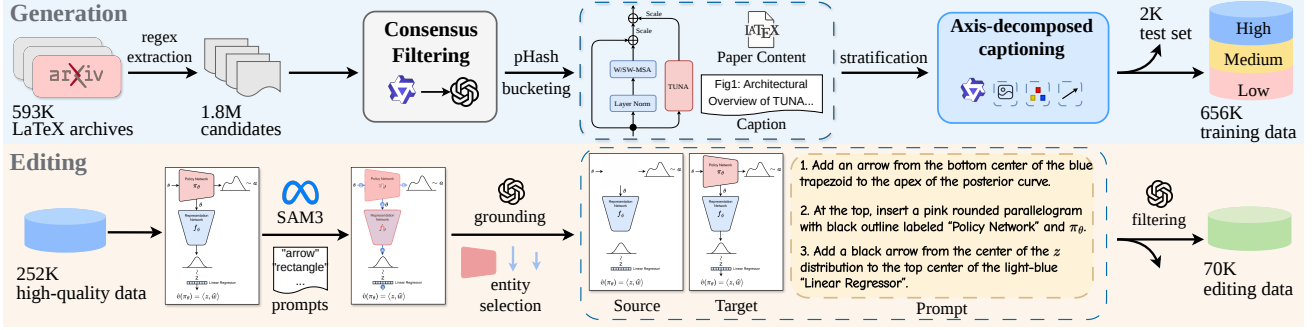}
    \Description{SciForma-700K data curation pipeline.}
    \vspace{-3mm}
    \caption{Overview of the SciFormaData-700K construction. \textbf{(Top:)} We curate 656K training samples through a pipeline of automated extraction, filtering, captioning, and stratification. \textbf{(Bottom:)} We synthesize 70K localized editing triplets by performing grounded entity modification on high-quality subsets.}
    \vspace{-3mm}
    \label{fig:data_pipeline}
\end{figure*}

\begin{table}[t]
\centering
\scriptsize
\vspace{-1mm}
\caption{Comparison of diagram benchmarks. \textbf{Elem.}: element verification; \textbf{Axes}: independent scores; \textbf{Diff.}: difficulty tiers; \textbf{Sev.}: error severity.}
\vspace{-3mm}
\label{tab:bench_comparison}
\setlength{\tabcolsep}{1.8pt}
\begin{tabular}{llccccc}
\toprule
Benchmark & Paradigm & Scale & Elem. & Axes & Diff. & Sev. \\
\midrule
ProImage-Bench~\cite{ProImage-Bench} & rubric scoring & 1{,}660 & \xmark & \xmark & \xmark & \xmark \\
GENFIG1~\cite{GENFIG1} & VLM grading & 2{,}450 & \xmark & \xmark & \xmark & \xmark \\
SciFlow-Bench~\cite{SciFlow-Bench} & graph matching & 500 & \cmark & \xmark & \xmark & \xmark \\
AIBench~\cite{AIBench} & hierarchical VQA & 300 & \xmark & \cmark & \xmark & \xmark \\
PaperBanana~\cite{PaperBanana} & pairwise win-rate & 292 & \xmark & \xmark & \xmark & \xmark \\
\midrule
\textbf{SciFormaBench-2K} & \textbf{inventory checklist} & \textbf{2{,}000} & \cmark & \cmark & \cmark & \cmark \\
\bottomrule
\end{tabular}
\vspace{-4mm}
\end{table}

\section{SciForma}
\label{sec:method}

\subsection{Overview}
\label{sec:overview}

Figure~\ref{fig:overview} illustrates SciForma's multi-stage pipeline: data curation, SFT, M-DPO, and iterative refinement.


\subsection{SciFormaData-700K}
\label{sec:data:dataset}

Existing scientific-figure datasets provide only coarse captions. As illustrated in Figure~\ref{fig:data_pipeline}a, we collect 726K methodology diagrams from 593K arXiv \LaTeX{} sources (2015--2025). Our pipeline employs consensus filtering by dual VLM evaluators and pHash deduplication, preserving high-fidelity visual assets and paragraph-level context.

\textbf{SciFormaData-700K} comprises \textbf{656K} generation pairs and \textbf{70K} editing triplets, constructed with three-aligned principles:

\noindent\textbf{Axis-decomposed captions.} Each diagram is paired with a caption organized along the Component, Arrow, and Text axes. The captions are synthesized from Qwen3-VL visual analysis integrated with the original paper context, averaging $547$ tokens per sample.

\noindent\textbf{Topology-aware stratification.} We partition the dataset into Low (13.6\%), Medium (48.2\%), and High (38.2\%) complexity tiers based on the structural inventory to enable two-stage fine-tuning.

\noindent\textbf{Inventory-aligned editing triplets (70K).} To support localized corrections, we construct 70K axis-specific edit triplets. We leverage SAM3~\cite{SAM3} and inpainting to perform grounded additions or deletions of 1--3 attributes. All edits are verified by OpenAI-o3 used for scalable data production, under the same structural-inventory, covering major modifications in arrows (54\%) and shapes (46\%). Details and statistics are provided in Appendix.

\subsection{SciFormaBench-2K}
\label{sec:bench:splits}

We hold out 2,000 candidates and stratify them by complexity: \textbf{Simple} (500) tests single-path block diagrams, \textbf{Medium} (900) introduces multi-branch flows with moderate topological complexity, and \textbf{Hard} (600) presents dense pipeline architectures with subfigures, cross-connections, and mathematical formulas. Each pairs with a human-verified inventory (Section~\ref{sec:inventory}), averaging 10.9 components, 10.2 arrows, and 6.6 text labels, with 56.8\% requiring formula rendering.

Applying the structural inventory to each image yields scores. For leaderboard ranking, we use cutting-edge GPT-5.4 for advanced visual reasoning, reporting an overall metric $\bar{s}$ while maintaining the granular decomposition for structural diagnosis.


Table~\ref{tab:bench_comparison} positions SciFormaBench-2K within the broader landscape. Existing benchmarks either rely on holistic scores~\cite{ProImage-Bench} or preferences~\cite{PaperBanana}, which conflate distinct structural errors and obscure failure modes. While SciFlow-Bench~\cite{SciFlow-Bench} and AIBench~\cite{AIBench} introduce finer granularities, they lack the combined axis-level decomposition and element-level attribution. Our benchmark focuses on verifying \textit{structural fidelity} in these dimensions, establishing a rigorous foundation for both verifiable evaluation and fine-grained alignment.

\subsection{Supervised Fine-Tuning (SFT)}
\label{sec:sft}
We employ a two-stage SFT strategy to adapt the base model to the structural constraints of scientific diagrams while maintaining flexibility across diverse aspect ratios. All stages utilize aspect-ratio bucketing to ensure spatial homogeneity within batches.

\noindent\textbf{Stage1: Structural adaptation.}
We first adapt FLUX.2-klein-base-9B~\cite{rename} to the fundamental diagram formation—layout, text rendering, and connectivity—using the 656K full training set at mid-resolution (${\sim}$768,px) with aspect-ratio bucketing.

\noindent\textbf{Stage2: Joint Generation and editing.}
To enable fine-grained structural editing capabilities, we co-train on high-quality generation samples and localized editing triplets at ${\sim}$1024,px. For editing, we concatenate source and target latents along the sequence dimension with a RoPE temporal offset ($T_{\text{target}}{=}0$, $T_{\text{source}}{=}10$), supervising only target tokens. This unified design allows a single architecture to handle both tasks without structural modifications.

This process finally leads to \textbf{SciForma-Base}. While proficient in general layout, it still exhibits residual structural errors, motivating the subsequent M-DPO stage to produce the final \textbf{SciForma-9B}.



\subsection{Multi-Dimensional Conjunctive Preference Optimization}
\label{sec:mdpo}

While SFT captures general diagram layout, it does not guarantee element correctness. One mistaken element still invalidates the whole figure. Although preference optimization like DPO~\cite{DPO} seems a natural next step, a global binary preference is too coarse: two rejected samples may fail for entirely orthogonal reasons, e.g., missing a directional arrow versus corrupting a formula subscript. Treating these as equivalent negative examples obscures the specific failure axes and provides no targeted correction signal.

To enforce conjunctive correctness across all axes, we propose M-DPO. Unlike standard post-training methods that collapse orthogonal structural axes into a single scalar reward, M-DPO maintains axis-specific optimization signals through two coupled design choices. First, dimension-anchored preference construction (Section~\ref{sec:mdpo:data}) isolates single-axis failures into targeted winner--loser pairs. Second, a conjunctive preference objective (Section~\ref{sec:mdpo:loss}), derived from a multi-way Bradley--Terry model, ensures the loss vanishes only when the model simultaneously prefers the winner across all structural dimensions.

\subsubsection{Dimension-Anchored Preference Construction}
\label{sec:mdpo:data}

Methodology-diagram failures are inherently axis-specific. Unlike standard scalar DPO, which collapses candidate pools into a single gradient direction, M-DPO captures these fine-grained failure modes. We rollout $K{=}12$ candidates per prompt,  and score them along $D{=}3$ structural axes using Qwen3-VL for scalable labeling. From this pool, the overall top candidate is selected as the shared winner $y^+$, and select per-axis losers $y^-_d$ that are weak on axis $d$ while remaining competitive on the others. Axes with no candidate variance are excluded from that prompt's loss. Unlike independent multi-axis DPO, sharing a single winner avoids gradient conflicts during optimization.

\subsubsection{Conjunctive Preference Objective}
\label{sec:mdpo:loss}
All $1{+}D$ images in a tuple share the same sampled noise $\varepsilon$ and timestep $t$ to eliminate noise-induced variance~\cite{D3PO}. Let $L_\theta(x) = \lVert v_\theta(x_t, t) - (\varepsilon{-}x) \rVert^2$ denote the per-sample flow-matching loss. The implicit reward logit for axis $d$ is
\begin{equation}
\label{eq:logit}
    \Delta_d = \bigl(L_{\mathrm{ref}}^{y^+} - L_\theta^{y^+}\bigr) - \bigl(L_{\mathrm{ref}}^{y^-_d} - L_\theta^{y^-_d}\bigr),
\end{equation}
measuring how much the policy prefers the winner $y^+$ over the axis-$d$ loser $y^-_d$, relative to the frozen reference model.

A natural baseline is the mean of $D$ independent DPO losses:
\begin{equation}
\label{eq:mean_dpo}
  \mathcal{L}_{\mathrm{mean\text{-}DPO}}
  = \frac{1}{D}\sum_{d=1}^{D} -\log\sigma(\beta\Delta_d).
\end{equation}
Though multi-axes, this baseline remains compensatory: when $D{-}1$ axes are already satisfied ($\Delta_d \gg 0$), the gradient on the single failing axis is diluted by the $1/D$ factor, and optimization can stall with one dimension still broken.
To enforce conjunctive correctness, we couple all axis-specific negatives through a single objective derived from a multi-way Bradley--Terry model:
\begin{equation}
\label{eq:mdpo}
  \mathcal{L}_{\mathrm{M\text{-}DPO}}
  = \log\Bigl(1 + {\textstyle\sum}_{d=1}^{D}\exp\bigl({-}\beta\,\Delta_d\bigr)\Bigr).
\end{equation}

This is the negative log-likelihood of a $(1{+}D)$-way comparison where the winner must beat every axis-specific loser. As a smooth relaxation of a hard-AND constraint, $\mathcal{L}_{\mathrm{M\text{-}DPO}} \to 0$ only when $\Delta_d > 0$ for all $d$; when $D{=}1$, Equation~\ref{eq:mdpo} reduces to standard DPO.

\paragraph{Adaptive gradient reweighting.}
Differentiating Equation~\ref{eq:mdpo} assigns each axis an effective gradient weight
\begin{equation}
\label{eq:grad_weight}
  w_d = \frac{\exp({-}\beta\,\Delta_d)}{1 + \sum_{d'}\exp({-}\beta\,\Delta_{d'})},
\end{equation}
a softmax over performance deficits. When the policy still prefers the loser on axis~$d$ ($\Delta_d < 0$), $w_d$ is large and the gradient concentrates on that axis; when all axes are satisfied, every $w_d \to 0$ and the loss vanishes. This adaptive reweighting routes the strongest learning signal to the most deficient axis without any hand-tuned schedule. The objective can also be viewed as an InfoNCE loss whose negatives are deliberately mined structural failure modes rather than random distractors. The full gradient derivation appears in Appendix.

\begin{figure}
    \centering
    \Description{M-DPO algorithm detail.}
    \includegraphics[width=0.98\linewidth]{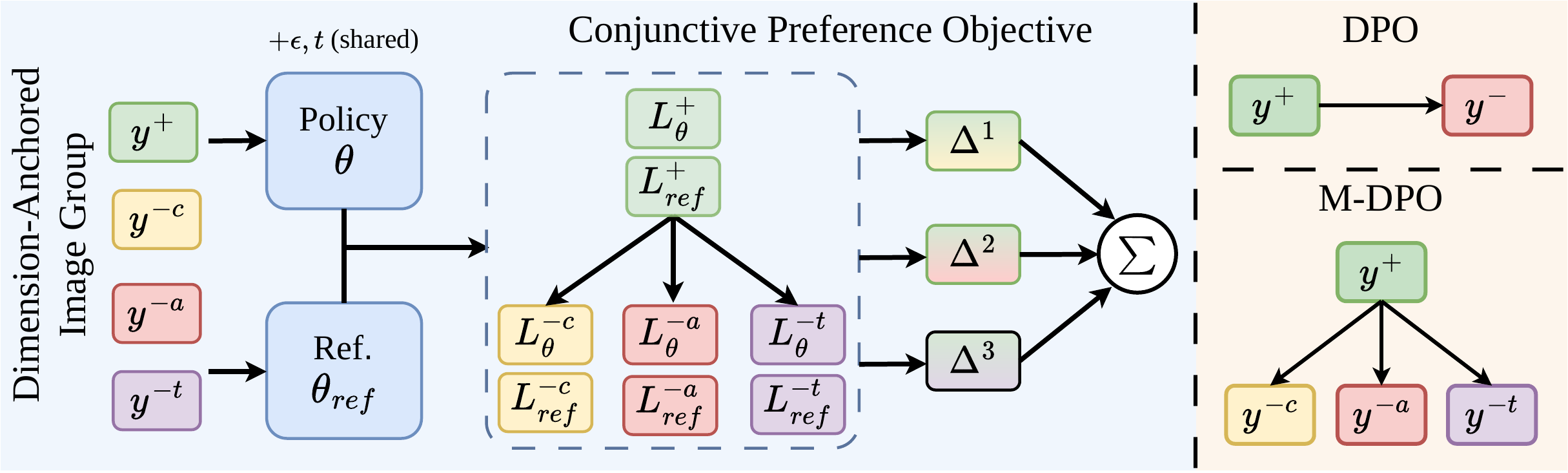}
    \vspace{-4mm}
    \caption{M-DPO Algorithm Overview. M-DPO computes axis-specific margins ($\Delta_d$) by pairing a winner with losers anchored to specific structural dimensions. These margins are aggregated via a conjunctive objective to prioritize optimization on the most deficient axes.}
    \vspace{-5mm}
    \label{fig:figure4}
\end{figure}

\begin{table*}[t!]
\centering
\caption{Left: SciFormaBench-2K (C/A/T = Component/Arrow/Text, S/M/H = Simple/Medium/Hard). Right: AIBench (Comp = Component, Topo = Topology, Phase = Phase Architecture, Sem = Semantics, Aes = Aesthetics/UniPercept). Color coding in the left panel: \colorbox{blue!10}{proprietary}, \colorbox{yellow!10}{open-source}, \colorbox{green!10}{ours}.}
\label{tab:sciforma_aibench_joint}
\vspace{-3mm}
\small
\begin{minipage}[t]{0.54\textwidth}
\centering
\textbf{(a) SciFormaBench-2K}\vspace{1mm}
\resizebox{\linewidth}{!}{%
\begin{tabular}{lccccccc}
\toprule
Method & Average & Simple & Medium & Hard & C & A & T \\
\midrule
\rowcolor{blue!10} GPT-Image-2 & 85.62 & 88.41 & 85.66 & 83.26 & 83.34 & 89.61 & 83.53 \\
\rowcolor{blue!10} Nano Banana Pro & 81.34 & 85.30 & 81.70 & 77.50 & 81.10 & 83.60 & 78.70 \\
\rowcolor{green!10} SciForma-9B + Edit & 72.40 & 79.82 & 72.59 & 66.01 & 76.70 & 69.91 & 70.14 \\
\rowcolor{green!10} SciForma-9B & \textbf{69.51} & \textbf{77.20} & \textbf{69.69} & \textbf{62.86} & \textbf{74.49} & \textbf{66.46} & \textbf{67.00} \\
\rowcolor{blue!10} GPT-Image-1.5 & 68.96 & 75.50 & 69.60 & 62.60 & 75.70 & 62.50 & 68.20 \\
\rowcolor{green!10} SciForma-Base & 67.59 & 76.01 & 67.43 & 60.83 & 73.52 & 64.64 & 63.84 \\
\rowcolor{blue!10} Wan2.7-Image & 64.71 & 73.90 & 65.90 & 55.30 & 71.90 & 60.30 & 61.10 \\
\rowcolor{yellow!10} SenseNova-U1-8B & 51.14 & 59.57 & 51.86 & 43.02 & 61.34 & 50.16 & 40.37 \\
\rowcolor{yellow!10} FLUX.2-dev 32B & 48.81 & 57.60 & 49.30 & 40.80 & 62.50 & 38.60 & 44.60 \\
\rowcolor{yellow!10} Qwen-Image-2512 & 48.73 & 57.50 & 49.90 & 39.60 & 61.20 & 46.60 & 36.80 \\
\rowcolor{yellow!10} Z-Image & 48.50 & 57.20 & 49.70 & 39.40 & 58.30 & 46.90 & 38.50 \\
\rowcolor{yellow!10} FLUX.2-klein-base-9B & 33.87 & 42.80 & 34.20 & 26.00 & 51.50 & 25.20 & 23.60 \\
\rowcolor{yellow!10} FLUX.1-dev & 20.64 & 26.40 & 20.70 & 15.80 & 40.60 & 11.20 & 8.50 \\
\rowcolor{yellow!10} Bagel 7B & 16.31 & 18.40 & 16.20 & 14.70 & 35.50 & 9.50 & 2.20 \\
\bottomrule
\end{tabular}%
}
\end{minipage}\hfill
\begin{minipage}[t]{0.445\textwidth}
\centering
\textbf{(b) AIBench}\vspace{1mm}
\resizebox{\linewidth}{!}{%
\begin{tabular}{lcccccc}
\toprule
Method & Score & Comp & Topo & Phase & Sem & Aes \\
\midrule
\rowcolor{blue!10} GPT-Image-2 & 80.27 & 90.69 & 81.77 & 84.18 & 90.24 & 54.49\\
\rowcolor{blue!10} Nano Banana Pro & 77.77 & 87.80 & 74.81 & 82.67 & 88.54 & 55.04 \\
\rowcolor{blue!10} Seedream~5.0 & 73.23 & 82.93 & 72.81 & 72.10 & 86.53 & 51.78 \\
\rowcolor{green!10} \textbf{SciForma-9B + Edit} & \textbf{70.62} & 77.86 & 64.47 & 75.45 & 81.20 & 54.10 \\
\rowcolor{green!10} \textbf{SciForma-9B} & \textbf{70.29} & 77.53 & 64.17 & 74.71 & 80.79 & 54.24 \\
Original Image & 70.09 & 82.65 & 57.98 & 79.57 & 79.13 & 51.11 \\
\rowcolor{yellow!10} SciForma-Base & 68.61 & 76.26 & 61.95 & 73.26 & 78.33 & 53.24 \\
\rowcolor{blue!10} Wan 2.6 & 65.84 & 68.60 & 56.11 & 72.56 & 80.43 & 51.50 \\
\rowcolor{blue!10} GPT-Image-1.5 & 61.62 & 66.23 & 50.87 & 55.95 & 77.55 & 57.50 \\
\rowcolor{blue!10} Seedream~4.5 & 59.68 & 67.89 & 52.42 & 48.14 & 74.47 & 55.48 \\
\rowcolor{yellow!10} Qwen-Image-2512 & 42.83 & 32.27 & 29.11 & 39.95 & 56.39 & 56.45 \\
\rowcolor{yellow!10} Z-Image & 41.62 & 25.61 & 34.42 & 57.41 & 54.00 & 36.65 \\
\rowcolor{yellow!10} FLUX.2-dev 32B & 37.94 & 23.40 & 24.84 & 52.55 & 46.52 & 42.40 \\
\rowcolor{yellow!10} Bagel 7B & 15.85 & 1.16 & 10.34 & 27.43 & 3.12 & 37.21 \\
\bottomrule
\end{tabular}%
}
\end{minipage}
\end{table*}

\subsection{Iterative Refinement}
\label{sec:editing_loop}

To address residual errors in high-density diagrams, we deploy \textbf{SciForma-9B} in an autonomous refinement loop, designed to monotonically improve \textit{structural fidelity} under explicit verification gates.

\noindent\textbf{Critic-guided localization.}
Guided by GPT-5.4 and the structural inventory (Section~\ref{sec:schema}), the system inspects the initial generation, localizes axis-specific defects (e.g., misrouted arrows, formula/text corruption), and ranks them by expected topological impact.

\noindent\textbf{Closed-loop editing.}
For each queued defect cluster, SciForma-9B performs localized inpainting on a semantic-block bounding box snapped to the $16{\times}16$ VAE grid. Conditioning on the global prompt, reference image, and region context preserves global intent while correcting local structure.

\noindent\textbf{Verification and rollback.}
Each edit proposes up to $K$ candidates scored under a fixed evaluator prompt and the same structural checklist; an edit is accepted only if it strictly improves the local score. A whole-image guard then recomputes global C/A/T scores, and if global consistency drops below the pre-edit baseline, the round is rolled back to the best known state.

\section{Experiments}
\label{sec:experiments}


We conduct experiments on three benchmarks: \textbf{SciFormaBench-2K} measures \textit{structural fidelity}. \textbf{AIBench}~\cite{AIBench} uses VQA to verify whether generated diagrams communicate research logic to the reader. \textbf{PaperBanana}~\cite{PaperBanana} evaluates whether SciForma can serve as a drop-in Visualizer in an agentic pipeline. Finally, ablation studies isolate the contributions of M-DPO's axis-decomposed pairs and conjunctive objective.
\subsection{Implementation Details}
\label{sec:exp_impl}

SciForma is initialized from FLUX.2-klein-base-9B~\cite{rename} and trained in \texttt{bfloat16} on $8{\times}$ B200 GPUs under DeepSpeed ZeRO-2. We apply AdamW (weight decay $10^{-2}$) and maintain float32 EMA ($\gamma{=}0.9999$)  weights for all reported checkpoints.

\textbf{SFT.} Stage~1 trains on 656K generation pairs at 768\,px for 140K steps (batch=16, LR=$10^{-5}$). Stage~2 initializes from the Stage-1 EMA and co-trains on 244K high-quality pairs with 70K editing triplets at ${\sim}$1024\,px for 90K steps.

\textbf{M-DPO.} From 50K prompts, we roll out $K{=}12$ candidates per prompt (four at 50 steps, eight at 25 steps), and score them via the structural-inventory with Qwen3-VL-8B-Instruct. Margin thresholds ($\delta_{\min}{=}0.25$, $\delta_{\max}{=}0.60$) and a winner gate ($\tau{=}0.70$) yield ${\sim}$16K valid $(1{+}D)$-tuples. Training runs for 4K steps on $4{\times}$ B200 GPUs (batch size=1, gradient accumulation=4, LR=$10^{-6}$, $\beta{=}2000$).

\textbf{Inference.} We use the Euler discrete scheduler with timestep shifting: 50 denoising steps, CFG scale 4.0. Unless noted otherwise, we evaluate all SciFormaBench-2K metrics with GPT-5.4. To verify that results are not evaluator-specific, we also provide a Qwen3-VL-8B-Instruct score in the Appendix.
\subsection{Results on SciFormaBench-2K}
\label{sec:exp_sciformabench}
On SciFormaBench-2K, we evaluate \textit{structural fidelity}. We benchmark 11 models at their default settings in 3 groups: \textbf{proprietary systems} (GPT-Image-2~\cite{GPTImage2}, Nano Banana Pro~\cite{NanoBananaPro}, GPT-Image-1.5~\cite{GPTImage1.5}, and Wan 2.7-Image~\cite{wan27image2024}); \textbf{open-source diffusion models} (FLUX.2-dev 32B~\cite{FLUX1dev}, Qwen-Image~\cite{QwenImage}, Z-Image~\cite{Z-Image}, FLUX.2-klein-base-9B~\cite{rename}, FLUX.1-dev~\cite{FLUX1dev}); and \textbf{omni models} (Bagel 7B~\cite{BAGEL}, SenseNova-U1-8B~\cite{SenseNovaU1}).

Proprietary models clearly surpass open-source models (Table~\ref{tab:sciforma_aibench_joint}a). \textbf{SciForma-Base} narrows this gap, lifting the overall performance of FLUX.2-klein-base-9B from 33.87 to 67.59, bringing it close to GPT-Image-1.5 (68.96).

\textbf{M-DPO corrects the remaining structural errors.} It lifts the overall performance of \textbf{SciForma-9B} to 69.51, surpassing GPT-Image-1.5. The largest gains appear on the two weakest axes: Arrow (+1.82) and Text (+3.16), confirming the adaptive-focus property in Section~\ref{sec:mdpo:loss}.

\textbf{Iterative Refinement} (SciForma-9B + edit) lifts the score to 72.40 (+2.89), with Arrow again gaining most. This shows that some topological errors need targeted repair at inference time.

\subsection{Results on AIBench}
\label{sec:exp_aibench}

AIBench~\cite{AIBench} tests whether a VLM can recover method logic from the generated diagram alone. Its VQA protocol evaluates four hierarchical levels (Component, Topology, Phase, and Semantics), plus an aesthetic score from UniPercept~\cite{UniPercept}. Because AIBench does not rely on our structural inventory, it also provides an independent check against circular evaluation. 

Following the benchmark’s standard protocol for diffusion models, we use a prompt rewriter to adapt raw \LaTeX{} into axis-decomposed captions, ensuring a level playing field for comparison. SciForma-9B reaches 70.29 (Table~\ref{tab:sciforma_aibench_joint}b), slightly edging out human-drawn originals (70.09) and widening the lead over GPT-Image-1.5 by 8.67 points. These results confirm that our structural improvements translate directly into superior readability for the end-reader.
 
The widest margin appears on Topology (+6.19 over originals), highlighting improved logical flow. Originals lead only in Component  (82.65 vs. 77.53). A likely reason is resolution: at 1024 px, dense diagrams with many small modules become harder to render clearly. 

\subsection{Agentic Integration on PaperBanana}
\label{sec:exp_agent}

PaperBanana~\cite{PaperBanana} tests whether SciForma can serve as a drop-in Visualizer inside an agentic pipeline. We evaluate on 292 samples with two adaptations on the original framework: a prompt rewriter condenses CSS-style scene descriptions ($\sim$5K chars) into in-domain prompts ($\sim$2.5K chars), and best-of-3 selection replaces the default critic-and-redraw loop. Gemini-3-Pro judges pairwise win rates against human-drawn originals across Faithfulness, Conciseness, Readability, and Aesthetics. 
\begin{table}[t]
\centering
\vspace{-3mm}
\caption{PaperBanana benchmark (292 samples, Gemini-3-Pro judge). Scores are pairwise win rates against human-drawn originals~(\%).}
\vspace{-3mm}
\label{tab:paperbanana}
\small
\resizebox{\linewidth}{!}{%
\begin{tabular}{lccccc}
\toprule
Method & Overall & Faith & Conc & Read & Aes \\
\midrule
PaperBanana + Nano Banana Pro & 60.2 & 45.8 & 80.7 & 51.4 & 72.1 \\
Nano Banana Pro & 43.2 & 43.0 & 43.5 & 38.5 & 65.5 \\
\midrule
\textbf{PaperBanana + SciForma (Ours)} & \textbf{30.7} & \textbf{34.5} & 37.7 & 31.2 & 52.6 \\
\midrule
PaperBanana + GPT-Image-1.5 & 19.0 & 16.0 & 65.0 & 33.0 & 56.0 \\
GPT-Image-1.5 & 11.5 & 4.5 & 37.5 & 30.0 & 37.0 \\
Paper2Any + Nano Banana Pro & 8.5 & 6.5 & 44.0 & 20.5 & 40.0 \\
\bottomrule
\vspace{-3mm}
\end{tabular}%
}
\vspace{-3mm}
\end{table}

\begin{figure*}
    \centering
    \Description{PaperBananaBench}
    \includegraphics[width=0.88\linewidth]{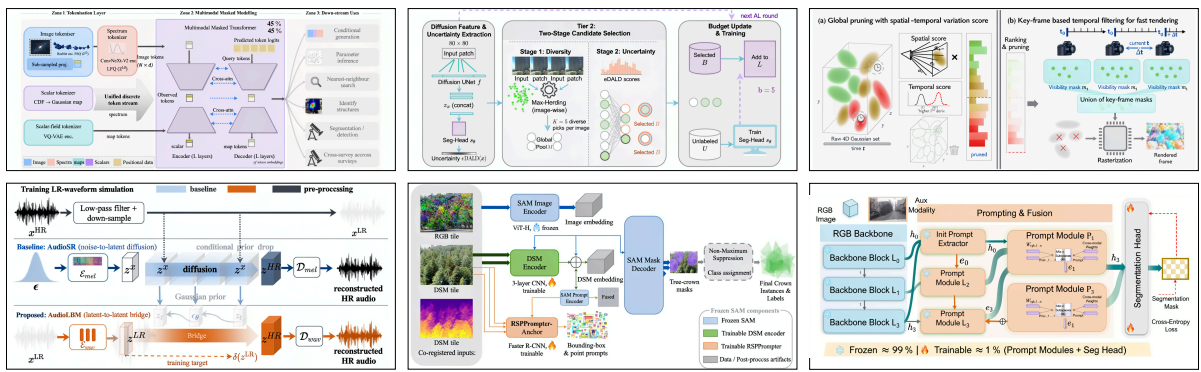} 
    \vspace{-3mm}
    \caption{Visualization of SciForma-9B on PaperBananaBench.}
    \label{fig:paperbanana}
    \vspace{-1mm}
\end{figure*}

SciForma achieves 30.7\% overall (Table~\ref{tab:paperbanana}), surpassing GPT-Image-1.5 and approaching Nano Banana Pro on Faithfulness (34.5\% vs.\ 45.8\%). The main bottleneck is Conciseness (37.7\% vs.\ 80.7\%): flow-matching models faithfully render every described element rather than selectively omitting details for cleaner layouts. Detailed visualizations are presented in Figure~\ref{fig:paperbanana}.

\subsection{Ablation Studies}
\label{sec:ablation}

We now deconstruct M-DPO by isolating the impact of its specific modules and optimization objectives. All ablations start from \textbf{SciForma-Base} and are scored on SciFormaBench-2K with GPT-5.4.

\begin{figure*}[t]
\centering
\Description{experiment of continue sft vs M-DPO}
\includegraphics[width=0.99\linewidth]{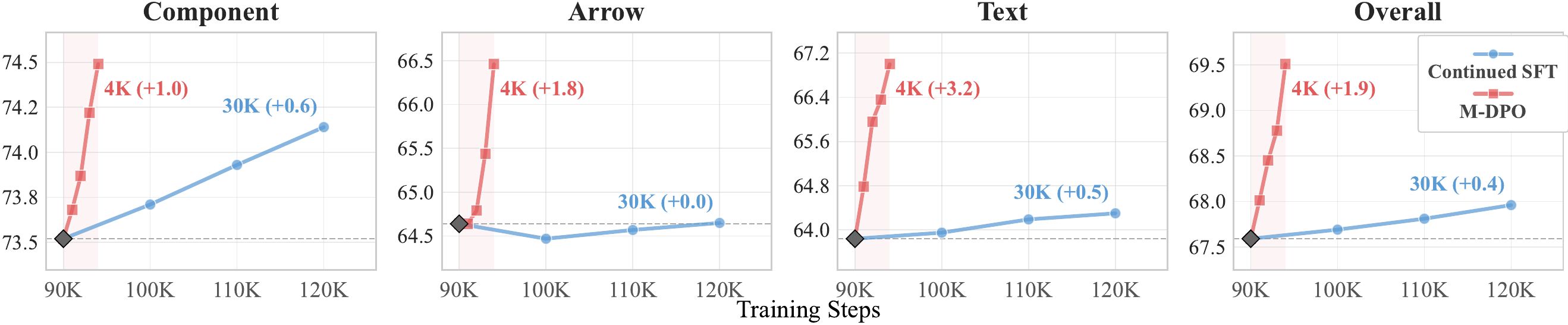}
\vspace{-3mm}
\caption{Training trajectories from the same 90K SFT checkpoint on SciFormaBench-2K (GPT-5.4). Red: M-DPO; Blue: continued SFT. M-DPO achieves larger gains in fewer steps, with strongest improvements on Text and Arrow.}
\label{fig:sft_vs_mdpo_traj}
\vspace{-3mm}
\end{figure*}

\paragraph{M-DPO breaks the SFT plateau.}
Figure~\ref{fig:sft_vs_mdpo_traj} plots two trajectories from the same starting point: 30K additional SFT steps versus 4K M-DPO steps. M-DPO achieves a larger average gain (+1.92 vs. +0.37) in one-seventh the training steps. The gains fall on the most deficient axes: Text +3.16 vs.\ +0.46, Arrow +1.82 vs.\ +0.01. Continued SFT sharpens axes already strong; M-DPO targets the remaining structural errors, exactly as the adaptive-focus mechanism predicts (Section~\ref{sec:mdpo:loss}).

\paragraph{Scalar-reward alternatives fail.} We test two well-established baselines: GDRO~\cite{GDRO}, which ranks candidates with a single overall reward via Plackett-Luce, and GRPO, which uses the same scalar in an online RL loop. The problem is the same for both: Component, Arrow, and Text collapse into one number, so the model is penalized without knowing which axis went wrong. A gain on one axis can quietly hide a regression on another. Table~\ref{tab:ablation_alt} confirms this: GDRO only shuffles errors between axes; GRPO makes things worse overall. The issue is not how hard the model trains, but what signal it trains on. Detailed axis-wise correlation analyses (see Appendix) lend further support to our statement.

\begin{table}[t]
\centering
\caption{Alternative post-training paradigms. Neither GDRO nor GRPO improves over the SFT baseline.}
\label{tab:ablation_alt}
\vspace{-3mm}
\small
\resizebox{\linewidth}{!}{%
\begin{tabular}{lcccc}
\toprule
Method & Average & C & A & T \\
\midrule
SciForma (SFT) & 67.59 & 73.52 & 64.64 & 63.84 \\
\midrule
+ GDRO  & 67.49~{\scriptsize \textcolor{red!80!black}{(-0.10)}} & 73.31~{\scriptsize \textcolor{red!80!black}{(-0.21)}} & 64.34~{\scriptsize \textcolor{red!80!black}{(-0.30)}} & 64.06~{\scriptsize \textcolor{green!60!black}{(+0.22)}} \\
+ GRPO  & 66.30~{\scriptsize \textcolor{red!80!black}{(-1.29)}} & 73.18~{\scriptsize \textcolor{red!80!black}{(-0.34)}} & 62.98~{\scriptsize \textcolor{red!80!black}{(-1.66)}} & 61.93~{\scriptsize \textcolor{red!80!black}{(-1.91)}} \\
\bottomrule
\end{tabular}
}
\vspace{-5mm}
\end{table}

\paragraph{The Necessity of Conjunctive Objectives.}

Table~\ref{tab:ablation_dpo2mdpo} presents an ablation study of various DPO learning objectives to validate our dimension-anchored construction strategy. We compare three baselines: (1) using global winner-loser pairs; (2) DPO (Pareto), which identifies winners and losers for each dimension independently, ensuring the winner's average score exceeds the loser's; and (3) M-DPO (mean) via Eq.~\ref{eq:mean_dpo}, serving as a compensatory baseline.

While Scalar DPO improves overall performance (+0.83), the gains are unevenly distributed across axes. While DPO (Pareto) provides fine-grained supervision, it introduces gradient conflicts that cause Text to regress, suggesting that multiple disparate winners complicate the optimization landscape. M-DPO (mean) resolves this with a shared winner, yet its $1/N$ scaling factor acts as a form of data augmentation that effectively shrinks the learning rate.In contrast, our conjunctive objective achieves the highest gain (+1.92) by concentrating gradients on the weakest axis.

This progression validates the theoretical framework addressed in Section~\ref{sec:mdpo:loss}: scalar preferences are compensatory, independent winners introduce conflicts, and shared winners resolve them. Conjunctive coupling focuses learning on the binding constraints. These ablations demonstrate that the primary bottleneck is not model capacity, but whether the supervision signal is decomposed along the same structural axes used to define diagram validity.

\paragraph{Comparison with peer multi-dimensional DPO methods.}
We further compare M-DPO against two recent multi-dimensional DPO variants: \textbf{CaPO}~\cite{lee2025calibrated}, which calibrates rewards for a Pareto-front DPO objective, and \textbf{MCDPO}~\cite{jang2025multi}, which conditions the DPO reward on per-axis preference vectors. Table~\ref{tab:ablation_peer_dpo} shows that neither method achieves balanced gains: both cause the Text axis to 
regress below the SFT baseline (CaPO $-1.34$, MCDPO $-1.85$), while M-DPO's conjunctive coupling lifts every axis with the largest gain on Text ($+3.16$). This confirms that M-DPO's improvement stems from its conjunctive formulation, rather than from simply being a multi-axis DPO variant.

\begin{table}[t]
\centering
\vspace{-4mm}
\caption{Ablation from scalar DPO to M-DPO on SciFormaBench-2K (GPT-5.4). Pair construction and loss aggregation are introduced step by step.}
\vspace{-4mm}
\label{tab:ablation_dpo2mdpo}
\small
\resizebox{\linewidth}{!}{%
\begin{tabular}{lcccc}
\toprule
Method & Average & C & A & T \\
\midrule
SciForma (SFT) & 67.59 & 73.52 & 64.64 & 63.84 \\
\midrule
+ DPO & 68.42~{\scriptsize \textcolor{green!60!black}{(+0.83)}} & 74.14~{\scriptsize \textcolor{green!60!black}{(+0.62)}} & 65.32~{\scriptsize \textcolor{green!60!black}{(+0.68)}} & 65.08~{\scriptsize \textcolor{green!60!black}{(+1.24)}} \\
+ DPO (Pareto) & 68.55~{\scriptsize \textcolor{green!60!black}{(+0.96)}} & 74.48~{\scriptsize \textcolor{green!60!black}{(+0.96)}} & 65.57~{\scriptsize \textcolor{green!60!black}{(+0.93)}} & 64.83~{\scriptsize \textcolor{green!60!black}{(+0.99)}} \\
\midrule
+ M-DPO (mean) & 69.31~{\scriptsize \textcolor{green!60!black}{(+1.72)}} & 74.54~{\scriptsize \textcolor{green!60!black}{(+1.02)}} & 66.28~{\scriptsize \textcolor{green!60!black}{(+1.64)}} & 66.47~{\scriptsize \textcolor{green!60!black}{(+2.63)}} \\
+ M-DPO & \textbf{69.51}~{\scriptsize \textcolor{green!60!black}{(+1.92)}} & \textbf{74.49}~{\scriptsize \textcolor{green!60!black}{(+0.97)}} & \textbf{66.46}~{\scriptsize \textcolor{green!60!black}{(+1.82)}} & \textbf{67.00}~{\scriptsize \textcolor{green!60!black}{(+3.16)}} \\
\bottomrule
\end{tabular}
}
\end{table}


\begin{table}[t]
\centering
\vspace{-2mm}
\caption{Comparison with peer multi-dimensional DPO methods on SciFormaBench-2K (GPT-5.4). Both CaPO and MCDPO cause Text-axis regression; only M-DPO's conjunctive coupling lifts every axis.}
\vspace{-3mm}
\label{tab:ablation_peer_dpo}
\small
\resizebox{\linewidth}{!}{%
\begin{tabular}{lcccc}
\toprule
Method & Average & C & A & T \\
\midrule
SciForma (SFT) & 67.59 & 73.52 & 64.64 & 63.84 \\
\midrule
+ CaPO~\cite{lee2025calibrated}  & 67.68~{\scriptsize \textcolor{green!60!black}{(+0.09)}} & 74.14~{\scriptsize \textcolor{green!60!black}{(+0.62)}} & 65.38~{\scriptsize \textcolor{green!60!black}{(+0.74)}} & 62.50~{\scriptsize \textcolor{red!80!black}{(-1.34)}} \\
+ MCDPO~\cite{jang2025multi} & 67.21~{\scriptsize \textcolor{red!80!black}{(-0.38)}} & 74.03~{\scriptsize \textcolor{green!60!black}{(+0.51)}} & 64.56~{\scriptsize \textcolor{red!80!black}{(-0.08)}} & 61.99~{\scriptsize \textcolor{red!80!black}{(-1.85)}} \\
\midrule
+ M-DPO (ours) & \textbf{69.51}~{\scriptsize \textcolor{green!60!black}{(+1.92)}} & \textbf{74.49}~{\scriptsize \textcolor{green!60!black}{(+0.97)}} & \textbf{66.46}~{\scriptsize \textcolor{green!60!black}{(+1.82)}} & \textbf{67.00}~{\scriptsize \textcolor{green!60!black}{(+3.16)}} \\
\bottomrule
\end{tabular}
}
\vspace{-3mm}
\end{table}

\subsection{Qualitative Results}
\label{sec:qualitative}

\begin{figure*}
    \centering
    \begin{subfigure}[b]{0.9\linewidth}
        \centering
        \includegraphics[width=\linewidth]{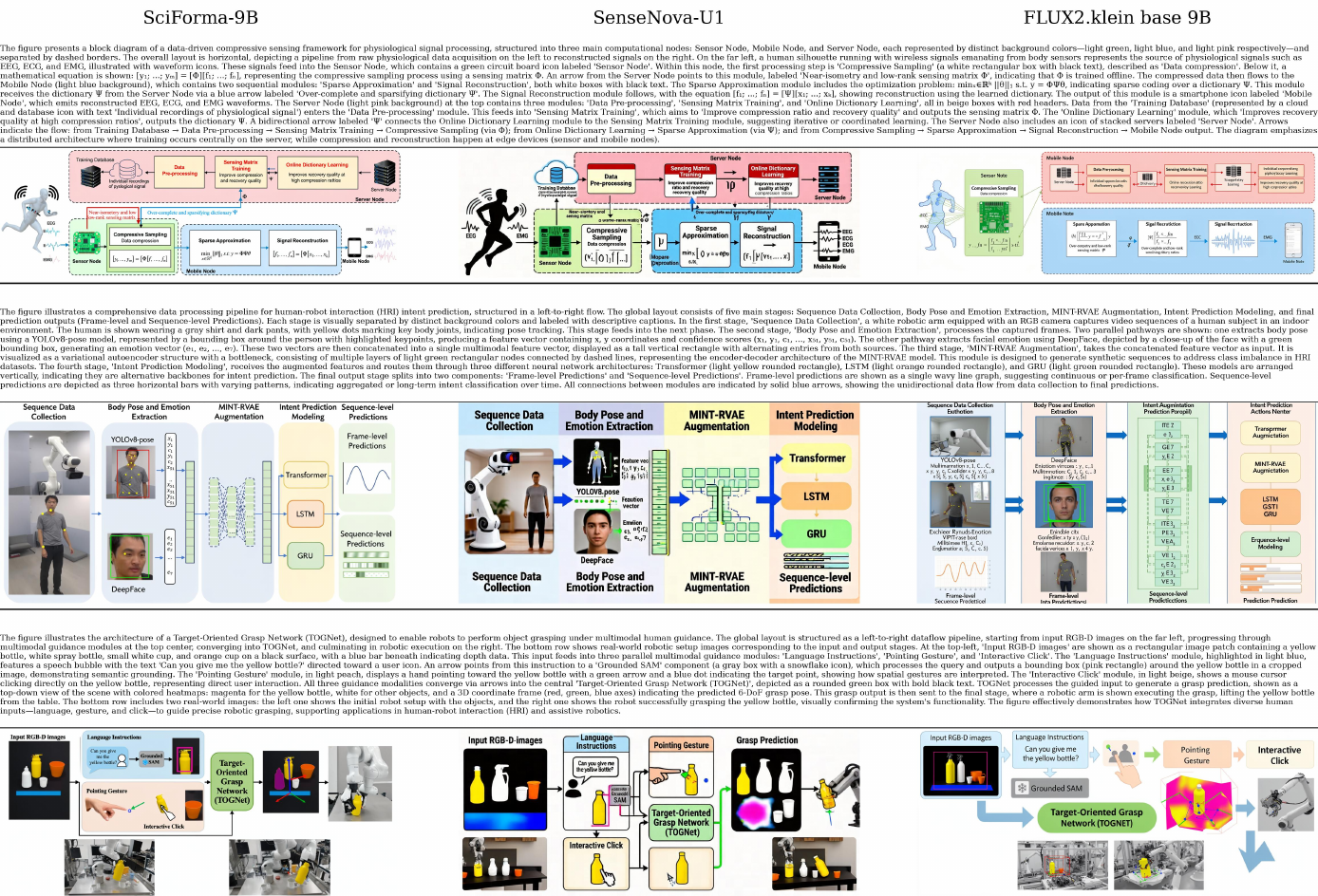}
        \caption{\textbf{Qualitative results of SciForma-9B against open-source models: SenseNova-U1 and FLUX.2-Klein-base-9B.}}
        \label{fig:sciforma_vs_sensenova}
    \end{subfigure}
    
    
    \begin{subfigure}[b]{0.9\linewidth}
        \centering
        \includegraphics[width=\linewidth]{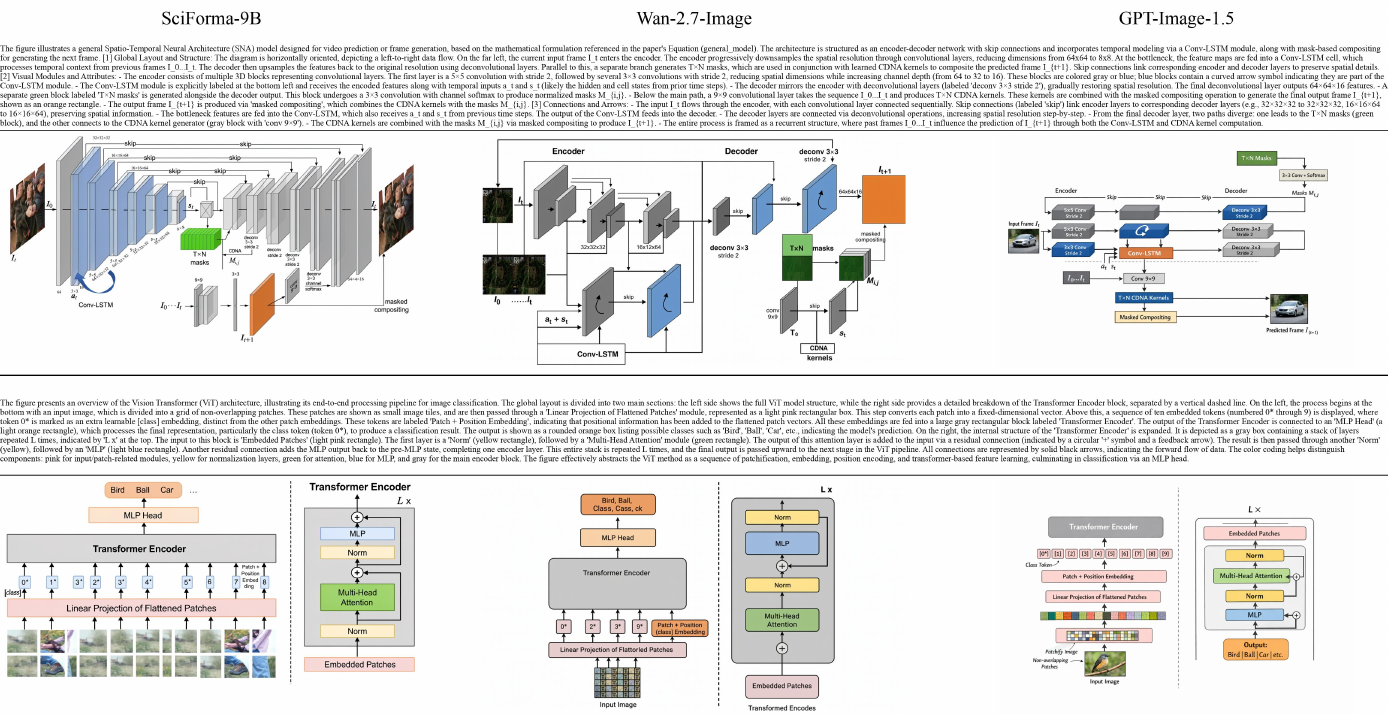}
        \caption{\textbf{Qualitative results of SciForma-9B against proprietary models: Wan2.7-Image and GPT-Image-1.5.}}
        \label{fig:sciforma_vs_wan}
    \end{subfigure}
    
    \caption{SciForma-9B generates high-fidelity pipeline figures with superior structural integrity and precise text rendering.}
    \Description{compare with wan etc}
    \label{fig:sciforma_comparisons}
\end{figure*}

Figures~\ref{fig:sciforma_vs_sensenova} and~\ref{fig:sciforma_vs_wan} compare SciForma-9B with open-source models (SenseNova-U1, FLUX.2-klein-base-9B) and proprietary models (Wan2.7-Image, GPT-Image-1.5). SciForma-9B generates structured diagrams with accurate text and logical spatial arrangement.

\begin{figure*}
    \centering
    \Description{Iterative Refinement}
    \includegraphics[width=0.92\linewidth]{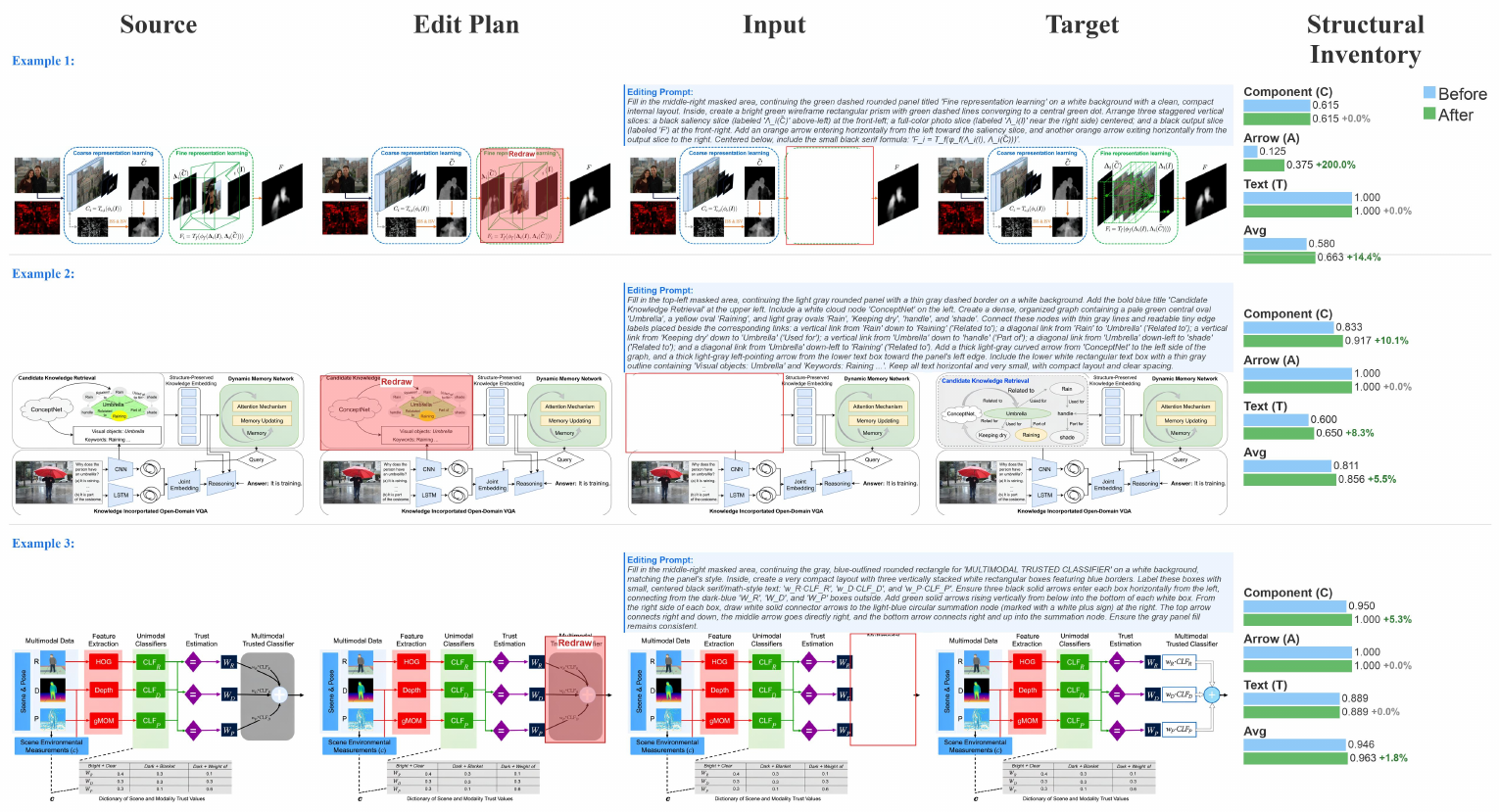} 
    \vspace{-3mm}
    \caption{Visualization of the closed-loop iterative refinement process.}
    \label{fig:editing}
    \vspace{-1mm}
\end{figure*}

Figure~\ref{fig:editing} shows the iterative refinement process. The structural inventory drives a closed-loop editing framework. Each round corrects distorted shapes, misaligned arrows, and inaccurate text. Candidate edits are evaluated against the structural checklist; if global consistency drops, the system reverts to the previous state. This strict verification-and-rollback mechanism enables fine-grained, multi-round editing while preserving overall generation quality.

\begin{figure*}
    \centering
    \Description{Visual Improvements introduced by M-DPO}
    \includegraphics[width=0.88\linewidth]{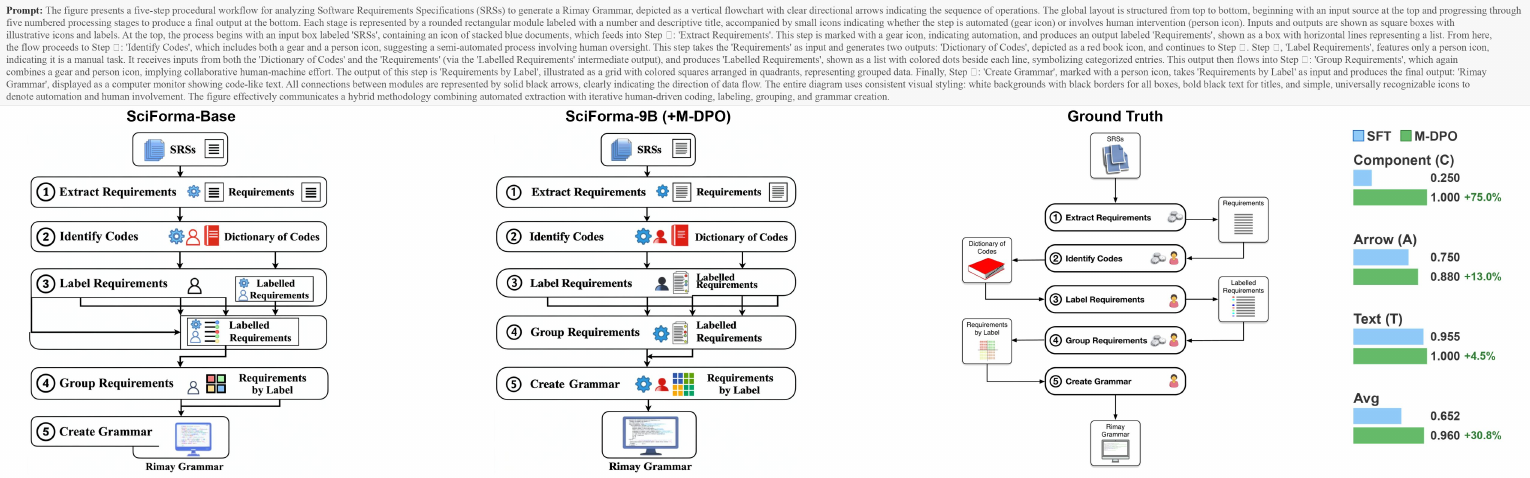}
    \vspace{-1mm}
    \caption{Visual improvements introduced by Multi-Dimensional Conjunctive Preference Optimization.}
    \label{fig:M-DPO}
\end{figure*}

Figure~\ref{fig:M-DPO} illustrates the visual improvements of M-DPO. Conjunctive optimization alleviates component misrendering and arrow distortion. The resulting diagrams show cleaner, more coherent topologies that closely align with the ground truth.

\subsection{User Study}
\label{sec:user_study}

\begin{figure}[t]
\centering
\vspace{-4mm}
\Description{User study.}
\includegraphics[width=0.99\linewidth]{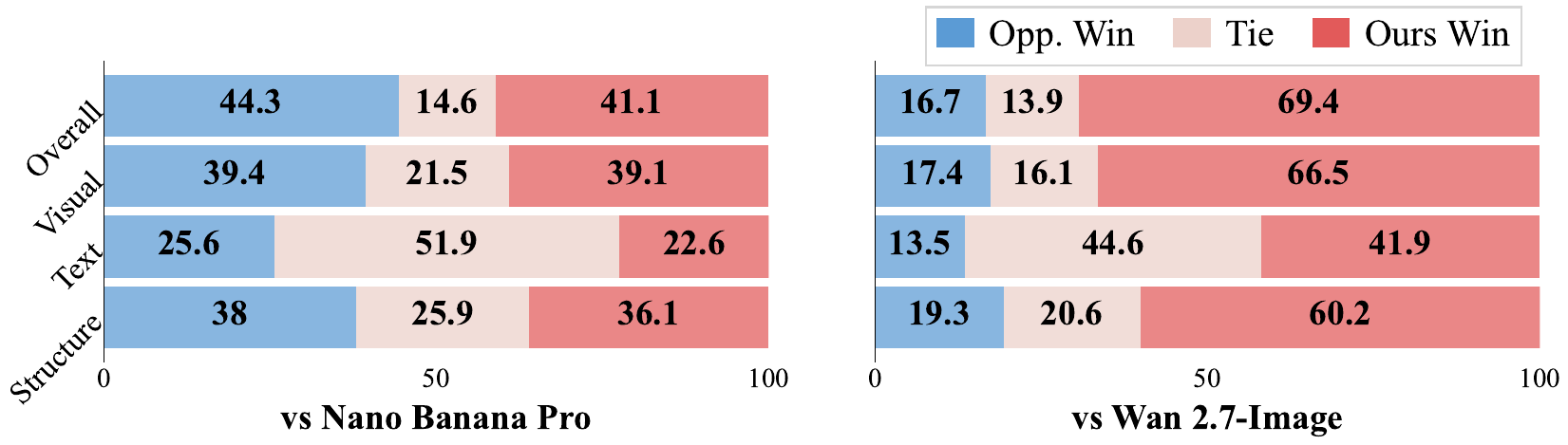}
\vspace{-4mm}
\caption{\textbf{User study.}
Win/Tie/Lose percentages for SciForma-9B against Nano Banana Pro and Wan2.7-Image across four evaluation dimensions.}
\label{fig:user_study}
\vspace{-3mm}
\end{figure}

We further conduct a user study, comparing SciForma-9B with Wan2.7-Image and Nano Banana Pro. Thirty-six graduate students specialized in AI evaluated 30 samples, voting on Structure, Text, Visual Quality, and 
Overall Preference. Figure~\ref{fig:user_study} shows that SciForma-9B outperforms Wan2.7-Image in all metrics and rivals Nano Banana Pro. This confirms that our gains in \textit{structural fidelity} also translate into human-perceived quality. Furthermore, the Pearson correlation between per-sample SciFormaBench-2K scores and user-study preferences reaches $r=0.76$, confirming positive alignment between our automated metric and human judgment.

\section{Discussion and Limitations}
\label{sec:limitations}
Although SciForma achieves structure-faithful diagram generation, some limitations remain: the 1024 px resolution limits fine-grained rendering in dense diagrams; the evaluation-and-editing pipeline depends on a proprietary VLM; and in agentic use, structural inventories are auto-generated rather than human-verified, risking drift from user intent. These limitations point toward higher-resolution training, open-source VLM evaluators, and human-in-the-loop agentic frameworks as next opportunities.

We release SciForma to boost open image-based diagram research and invite the community to extend M-DPO's conjunctive design to per-axis tasks. To prevent misuse in paper fabrication, we advocate human verification and in-the-loop editing of all generated outputs.

\section{Conclusion}
\label{sec:conclusion}
SciForma makes scientific diagram generation structure-faithful. We formulate diagram quality through three verifiable axes: Component, Arrow, and Text, and enforce their joint correctness with M-DPO. Experiments show that SciForma-9B approaches proprietary-level structural fidelity and aesthetics with explicit structural constraints.



\section{Acknowledgement}
This work was supported by National Natural Science Foundation of China (Grant No.: 62372015), Leading Projects in Key Research Fields of Language Funded by the National Language Commission, Center For Chinese Font Design and Research, Key Laboratory of Intelligent Press Media Technology, and National Engineering Research Center of New Electronic Publishing Technologies.
\newpage

\clearpage
\setcounter{section}{0}               
\renewcommand{\thesection}{\Alph{section}}

\section*{Supplementary Material Overview}

\begin{itemize}
    \item \hyperref[app:dataset_details_main]{Appendix A: SciFormaData-700K Details}
    \item \hyperref[app:benchmark_details]{Appendix B: SciFormaBench-2K Details}
    \item \hyperref[app:mdpo_analysis]{Appendix C: M-DPO Theoretical Analysis}
    \item \hyperref[app:implementation]{Appendix D: Implementation Details}
    \item \hyperref[app:sciformabench_analysis]{Appendix E: SciFormaBench-2K In-depth Analysis}
    \item \hyperref[app:more_ablations]{Appendix F: More Ablations}
    \item \hyperref[app:qualitative_results]{Appendix G: Additional Qualitative Results}
    \item \hyperref[app:user_study_details]{Appendix H: User Study Details}
    \item \hyperref[app:all_prompts]{Appendix I: Prompt Templates}
\end{itemize}

\section{SciFormaData-700K Details}
\label{app:dataset_details_main}

\subsection{Dataset Construction}

\label{app:dataset_details}

SciFormaData-700K is assembled through a four-stage pipeline. The full prompt templates for all VLM-mediated steps are provided in Appendix~\ref{app:all_prompts}.

\subsubsection{Stage~1: Source Extraction}

We parsed \LaTeX{} sources from ~593K arXiv papers (covering 17 \texttt{cs.*} categories from 2015 to 2025). Direct \LaTeX{} extraction ensures high-fidelity assets by avoiding PDF parsing artifacts. We automatically identify \texttt{figure} environments to extract image assets, captions, and their corresponding textual references. An LLM-based parser resolves custom macros (e.g., \texttt{\textbackslash newcommand}) into structured metadata.

\subsubsection{Stage~2: Consensus Filtering}

We filter $\sim$1.8M candidates for methodology diagrams, discarding plots, tables, and artifacts. A dual-VLM consensus retains only figures independently classified as methodology targets by both models.

\paragraph{Duplicate Removal and Structural Integrity.} 
We deduplicate the dataset using Perceptual Hashing (pHash) structural fingerprints. Specifically, we utilize Perceptual Hashing (pHash) to compute structural fingerprints for the extracted diagrams. This allows us to strictly filter out structurally redundant instances---such as exact-duplicate renders, parsing artifacts, and highly overlapping diagrams---based on a strict Hamming distance threshold. This ensures topological diversity and prevents information collapse between training and benchmark sets.

\subsubsection{Stage~3: Captioning and Stratification}

Qwen3-VL generates axis-decomposed captions by integrating visual features with \LaTeX{} context. These descriptions decompose each diagram into the Component, Arrow, and Text (C/A/T) axes. Based on the extracted structural inventory complexity, we stratify the pool into \textsc{Low} (13.6\%), \textsc{Medium} (48.2\%), and \textsc{High} (38.2\%) complexity tiers. We hold out a 2,000-sample balanced benchmark (SciFormaBench-2K), yielding the final 656K generation training pairs.

\subsubsection{Stage~4: Editing Triplet Construction}

We derive 70K editing triplets from a ${\sim}$252K high-quality subset. We utilize SAM3~\cite{SAM3} to segment semantic regions, then randomly sample 1--3 components or arrows to pre-compute deterministic editing paths (addition, deletion, etc.). Edits are validated for visual integrity (artifact-free) and semantic grounding (attribute alignment). Invalid branches are discarded in a traceback scheme, yielding 70K high-purity edit pairs for Stage-2 SFT.

\begin{figure}[t]
    \centering
    \includegraphics[width=0.99\linewidth]{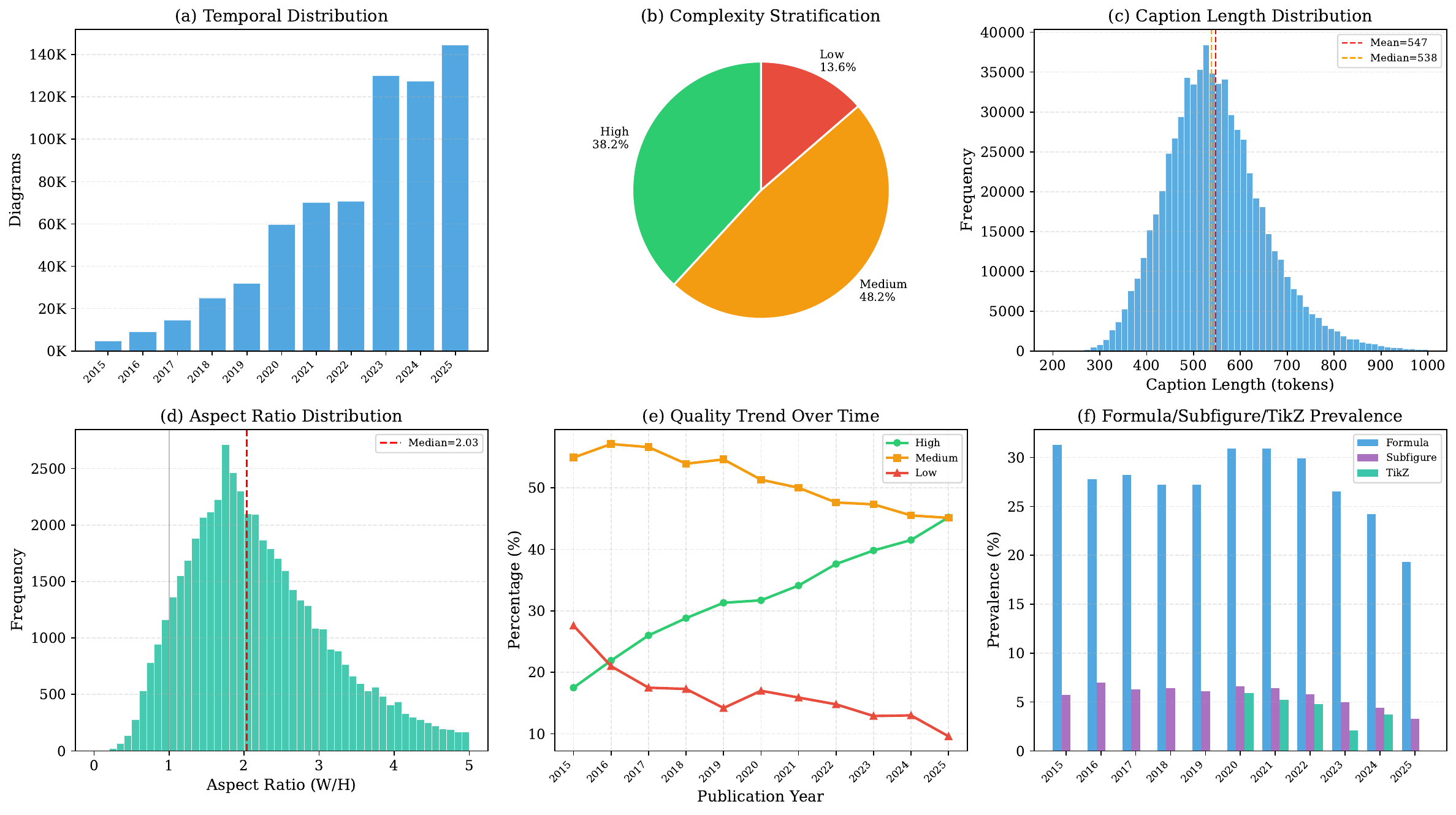}
    \vspace{-3mm}
    \caption{SciFormaData-700K statistics. From left to right: temporal growth of source papers and extracted figures (a), complexity stratification (b), caption token-length distribution (c), resolution and aspect-ratio distribution (d), quality trend over time (e), and prevalence of structural elements.}
    \label{fig:data_stats_combined}
\end{figure}

\begin{table}[ht]
    \centering
    \vspace{-3mm}
    \caption{Data construction funnel for SciFormaData-700K.}
    \vspace{-3mm}
    \label{tab:data_funnel}
    \small
    \renewcommand{\arraystretch}{1.2}
    \begin{tabular*}{\columnwidth}{@{\extracolsep{\fill}}lr}
        \toprule
        Stage & Count \\
        \midrule
        ArXiv paper IDs crawled (2015--2025) & 593{,}536 \\
        Raw candidate figures extracted & ${\sim}$1{,}800{,}000 \\
        \midrule
        \textbf{Generation Data Pathway} & \\
        After deduplication \& filtering & 658{,}000 \\
        Held-out benchmark (SciFormaBench-2K) & 2{,}000 \\
        \textbf{Final generation training set} & \textbf{656{,}000} \\
        \quad $\hookrightarrow$ \textsc{High} complexity (38.2\%) & 250{,}592 \\
        \quad $\hookrightarrow$ \textsc{Medium} complexity (48.2\%) & 316{,}192 \\
        \quad $\hookrightarrow$ \textsc{Low} complexity (13.6\%) & 89{,}216 \\
        \midrule
        \textbf{Editing Data Pathway} & \\
        High-quality candidates processed & 252{,}000 \\
        \textbf{Final verifiable editing triplets} & \textbf{70{,}000} \\
        \midrule
        \textbf{Total SciFormaData-700K Volume} & \textbf{726{,}000} \\
        \bottomrule
    \end{tabular*}
\end{table}

\subsection{Dataset Statistics and Analysis}
\label{app:data_stats}

The distributions of SciFormaData-700K are visualized in Figure~\ref{fig:data_stats_combined}, with the construction funnel detailed in Table~\ref{tab:data_funnel}.

\noindent\textbf{Topic coverage.} Source papers are drawn from 17 ArXiv Computer Science subcategories, including \texttt{cs.CV}, \texttt{cs.LG}, \texttt{cs.AI}, \texttt{cs.CL}, \texttt{cs.RO}, \texttt{cs.NE}, \texttt{cs.IR}, \texttt{cs.HC}, \texttt{cs.SE}, \texttt{cs.DC}, \texttt{cs.PL}, \texttt{cs.CG}, \texttt{cs.GR}, \texttt{cs.DB}, \texttt{cs.MM}, \texttt{cs.MA}, and \texttt{cs.SD}, thereby covering the major areas in which methodology diagrams frequently appear.

\noindent\textbf{Complexity and Quality.} A key strength is the professional complexity stratification (Figure~\ref{fig:data_stats_combined}b): the collection is predominantly composed of \textsc{High} (38.2\%) and \textsc{Medium} (48.2\%) samples. Over the past decade, the \textsc{High}-quality fraction has grown steadily, reaching 45.2\% in 2025 (Figure~\ref{fig:data_stats_combined}e), reflecting the increasing sophistication of modern research pipelines. This enables a curated curriculum using the ${\sim}$250K \textsc{High}-complexity subset for Stage-2 SFT.

\noindent\textbf{Information Density.} The dataset is curated for high-fidelity generation due to its dense annotations. As shown in Figure~\ref{fig:data_stats_combined}c, axis-decomposed captions feature a substantial median length of 538 tokens. These long prompts provide the precise structural detail required to reconstruct methodology logic from text alone.

\noindent\textbf{Structural Diversity.} Geometric and structural priors align with academic conventions: 89.5\% of diagrams are landscape with a median aspect ratio of 2.03 (Figure~\ref{fig:data_stats_combined}d). Furthermore, 25.9\% of diagrams contain embedded mathematical formulae and 2.6\% are TikZ-rendered (Figure~\ref{fig:data_stats_combined}f). Finally, 71.4\% of the 70K editing triplets involve 2–3 concurrent modifications, providing a rigorous signal for maintaining identity across visual transformations.
\section{SciFormaBench-2K Details}
\label{app:benchmark_details}

This section details the SciFormaBench-2K. We elaborate on the structural inventory extraction process (Section~\ref{app:schema_detail}) and define the specific error taxonomy used for scoring (Section~\ref{app:eval_taxonomy}).

\subsection{Structural Inventory}
\label{app:schema_detail}

We adopt Components (C), Arrows (A), and Text (T) as verifiable structural primitives for reproducible assessment, bypassing subjective aesthetic metrics. The inventory is extracted via joint prompt-image analysis using GPT-5.4 or direct \LaTeX{} parsing (Appendix~\ref{app:all_prompts}). During evaluation, a VLM performs a two-way comparison between the ground-truth and generated images against this C/A/T checklist to identify structural discrepancies.

\subsection{Evaluation Taxonomy}
\label{app:eval_taxonomy}

Table~\ref{tab:error_taxonomy} defines our error taxonomy and associated weights.

\paragraph{Severity Rationale.} Errors are weighted by their impact on structural logic: critical errors ($w_e = 1.0$) represent fundamental misrepresentations or missing elements, while moderate errors ($w_e = 0.5$) cover functional but defective elements (e.g., duplicates or semantic shifts). This allows capturing nuanced flaws without penalizing the primary topological skeleton.

\paragraph{Verification Protocol.} To ensure objectivity, evaluation ignores non-topological traits such as arrow curvature or routing paths. We decompose the evaluation into three axis-specific VLM calls (prompts in Appendix~\ref{app:eval_prompt_templates}) to prevent context exhaustion and maximize checking focus.

\begin{table}[t]
\centering
\caption{Error taxonomy for SciFormaBench-2K evaluation. Each error type is associated with a fixed severity and weight.}
\label{tab:error_taxonomy}
\scriptsize
\setlength{\tabcolsep}{3pt}
\begin{tabular}{llcp{4.8cm}}
\toprule
\textbf{Axis} & \textbf{Error Type} & \textbf{Severity} & \textbf{Description} \\
\midrule
\multirow{6}{*}{\textbf{C}} 
  & Missing        & critical  & Reference component is entirely absent. \\
  & Hallucinated   & critical  & Major element that clearly should not exist; random noise or obvious artifact. \\
  & Distorted      & critical  & Shape is severely distorted, blurred, or illegible. \\
  & Structural mismatch & moderate  & Exists at correct location but internal structure differs significantly. \\
  & Duplicate      & moderate  & Appears more times than in the reference. \\
  & Wrong          & moderate  & Present at correct location but depicts a clearly different concept. \\
\midrule
\multirow{4}{*}{\textbf{A}} 
  & Missing        & critical  & Reference connection is entirely absent. \\
  & Hallucinated   & critical  & Connections that clearly should not exist. \\
  & Position error & critical  & Arrow exists but start or end points are noticeably off. \\
  & Wrong          & moderate  & Arrow connects wrong components or points in a wrong direction. \\
\midrule
\multirow{5}{*}{\textbf{T}} 
  & Missing        & critical  & Reference label is entirely absent. \\
  & Garbled / unreadable & critical  & Character-level corruption: misspellings, scrambled characters, distorted glyphs. \\
  & Truncated      & critical  & Label is cut off or incomplete. \\
  & Wrong text     & moderate  & Fully legible but says something completely different from the reference. \\
  & Duplicated     & moderate  & Same label appears where it should not. \\
\bottomrule
\end{tabular}
\end{table}

\begin{figure}[t]
    \centering
    \Description{Illustration of the SciFormaBench-2K evaluation pipeline.}
    \includegraphics[width=0.99\linewidth]{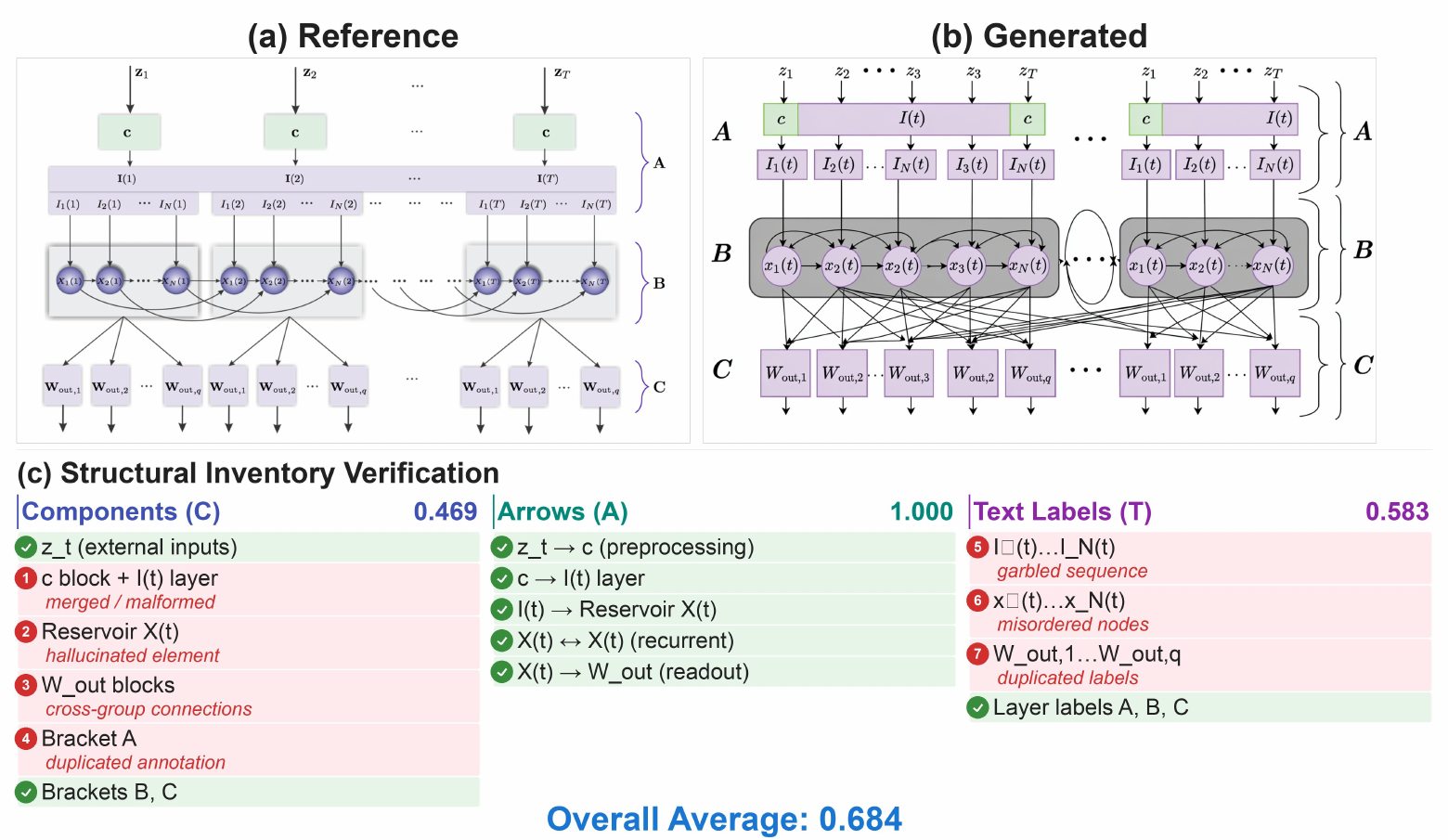}
    
    \vspace{-3mm}
    \caption{Illustration of structural inventory verification on a generated diagram. (a)Reference image. (b)Generated output. (c)Per-item pass/fail checklist across three evaluation dimensions.}
    \label{fig:BenchmarkEvaluationPlaceholder}
\end{figure}

As illustrated in Figure~\ref{fig:BenchmarkEvaluationPlaceholder}, this protocol provides actionable component-level feedback. The visualization details the exact format of the structural inventory and demonstrates how the system assigns an error type and severity score to each discrepancy.

\clearpage

\section{M-DPO Theoretical Analysis}
\label{app:mdpo_analysis}

This appendix provides the full derivation of the adaptive gradient-focus dynamics and establishes formal connections to the multi-way Bradley-Terry model and InfoNCE.

\subsection{Gradient Derivation}
\label{app:mdpo:gradient}

We restate the M-DPO objective:
\begin{equation}
  \mathcal{L}_{\mathrm{M\text{-}DPO}}
  = \log\Bigl(1 + {\textstyle\sum}_{d=1}^{D}\exp\bigl({-}\beta\,\Delta_d\bigr)\Bigr),
\end{equation}
where $\Delta_d = \bigl(L_{\mathrm{ref}}^{y^+} - L_\theta^{y^+}\bigr) - \bigl(L_{\mathrm{ref}}^{y^-_d} - L_\theta^{y^-_d}\bigr)$ is the implicit reward logit for axis $d$.

Since $L_{\mathrm{ref}}^{y^+}$ and $L_{\mathrm{ref}}^{y^-_d}$ are constants (the reference model is frozen), $\Delta_d$ depends on $\theta$ only through the policy losses $L_\theta^{y^+}$ and $L_\theta^{y^-_d}$. Differentiating with respect to $\theta$:

\begin{align}
\frac{\partial \mathcal{L}_{\mathrm{M\text{-}DPO}}}{\partial \theta}
&= \frac{\sum_{d=1}^{D} \exp({-}\beta\,\Delta_d)\cdot({-}\beta)\,\frac{\partial \Delta_d}{\partial \theta}}{1 + \sum_{d'=1}^{D}\exp({-}\beta\,\Delta_{d'})} \nonumber \\[4pt]
&= -\beta \sum_{d=1}^{D} w_d \,\frac{\partial \Delta_d}{\partial \theta},
\label{eq:grad_full}
\end{align}
where the \emph{effective gradient weight} for axis $d$ is
\begin{equation}
\boxed{\;w_d = \frac{\exp({-}\beta\,\Delta_d)}{1 + \sum_{d'=1}^{D}\exp({-}\beta\,\Delta_{d'})}\;}
\label{eq:grad_weight_app}
\end{equation}

Expanding the per-axis logit derivative:
\begin{equation}
\frac{\partial \Delta_d}{\partial \theta} = -\frac{\partial L_\theta^{y^+}}{\partial \theta} + \frac{\partial L_\theta^{y^-_d}}{\partial \theta},
\end{equation}
the full gradient becomes
\begin{equation}
\frac{\partial \mathcal{L}_{\mathrm{M\text{-}DPO}}}{\partial \theta}
= \beta \sum_{d=1}^{D} w_d \left(\frac{\partial L_\theta^{y^+}}{\partial \theta} - \frac{\partial L_\theta^{y^-_d}}{\partial \theta}\right).
\label{eq:grad_expanded}
\end{equation}

Each gradient step simultaneously decreases the flow-matching loss on the winner $y^+$ and increases it on every axis-anchored loser $y^-_d$, weighted by $w_d$.

\subsection{Adaptive Focus Property}
\label{app:mdpo:autofocus}

The weights $\{w_d\}_{d=1}^{D}$ in Equation~\ref{eq:grad_weight_app} form a softmax distribution over axes, parameterized by the negated logits $\{-\beta\Delta_d\}$. This yields the following adaptive behavior:

\begin{itemize}[leftmargin=*, topsep=2pt, itemsep=3pt, parsep=0pt]
    \item \textbf{Failing axis ($\Delta_d < 0$).} The policy still prefers the loser over the winner on axis $d$. Then $-\beta\Delta_d > 0$, making $\exp(-\beta\Delta_d) > 1$ and $w_d$ large.

    \item \textbf{Satisfied axis ($\Delta_d \gg 0$).} When the policy strongly prefers the winner, $\exp(-\beta\Delta_d) \approx 0$ and $w_d \approx 0$.
    \item \textbf{All axes satisfied ($\forall d\!:\, \Delta_d \gg 0$).} Under this circumstance, the denominator, $1 + \sum_{d'}\exp(-\beta\Delta_{d'}) \approx 1$, so $\sum_d w_d \approx 0$ and $\mathcal{L}_{\mathrm{M\text{-}DPO}} \to 0$. The loss vanishes and training halts, as desired.
\end{itemize}

The ``$+1$'' in Equation~\ref{eq:grad_weight_app}'s denominator acts as a no-update anchor: when all axes are satisfied, the gradient does not redistribute to some arbitrary axis but instead zeroes out entirely. This prevents the over-optimization pathology where a converged model is pushed to continue changing.

\noindent\textbf{Contrast with the mean baseline.} The per-axis mean objective $\frac{1}{D}\sum_d -\log\sigma(\beta\Delta_d)$ assigns a fixed $1/D$ weight to each axis regardless of its current performance. When $D{-}1$ axes have $\Delta_d \gg 0$, the gradient for the single failing axis is diluted by the $1/D$ factor, stalling optimization. M-DPO avoids this by dynamically concentrating weight on the bottleneck.

\noindent\textbf{Reduction to standard DPO ($D{=}1$).}
\label{app:mdpo:reduction}
When $D{=}1$ there is a single loser $y^-_1$ and the loss becomes $\mathcal{L}_{\mathrm{M\text{-}DPO}}\big|_{D=1} = \log\bigl(1 + e^{-\beta\,\Delta_1}\bigr) = -\log\sigma(\beta\,\Delta_1)$, using $\log(1{+}e^{-x})=-\log\sigma(x)$. This is exactly the standard DPO loss; M-DPO thus strictly generalizes scalar DPO to conjunctive multi-axis optimization for $D \ge 2$.

\subsection{Multi-Way Bradley-Terry Derivation}
\label{app:mdpo:bt}

We derive M-DPO from a multi-way Bradley-Terry preference model. Consider a comparison where the winner $y^+$ must be preferred over $D$ axis-anchored losers $\{y^-_d\}$ simultaneously. Under the Bradley-Terry framework, the probability that $y^+$ wins over all competitors is
\begin{equation}
P(y^+ \succ y^-_1, \ldots, y^-_D) = \frac{\exp(r^+)}{\exp(r^+) + \sum_{d=1}^{D}\exp(r^-_d)},
\end{equation}
where $r^+$ and $r^-_d$ are the implicit rewards of the winner and axis-$d$ loser respectively.

Taking the negative log-likelihood:
\begin{align}
-\log P &= -\log\frac{\exp(r^+)}{\exp(r^+) + \sum_d \exp(r^-_d)} \nonumber \\
&= \log\Bigl(1 + \sum_d \exp(r^-_d - r^+)\Bigr) \nonumber \\
&= \log\Bigl(1 + \sum_d \exp(-\beta\Delta_d)\Bigr),
\end{align}
where $\Delta_d = \frac{1}{\beta}(r^+ - r^-_d)$ is the implicit reward logit from DPO reparameterization.

The formulation makes conjunctive semantics explicit: training rewards the policy if and only if the winner is preferred over \emph{every} axis-specific loser simultaneously.

\begin{figure*}[t]
\centering
\includegraphics[width=\linewidth]{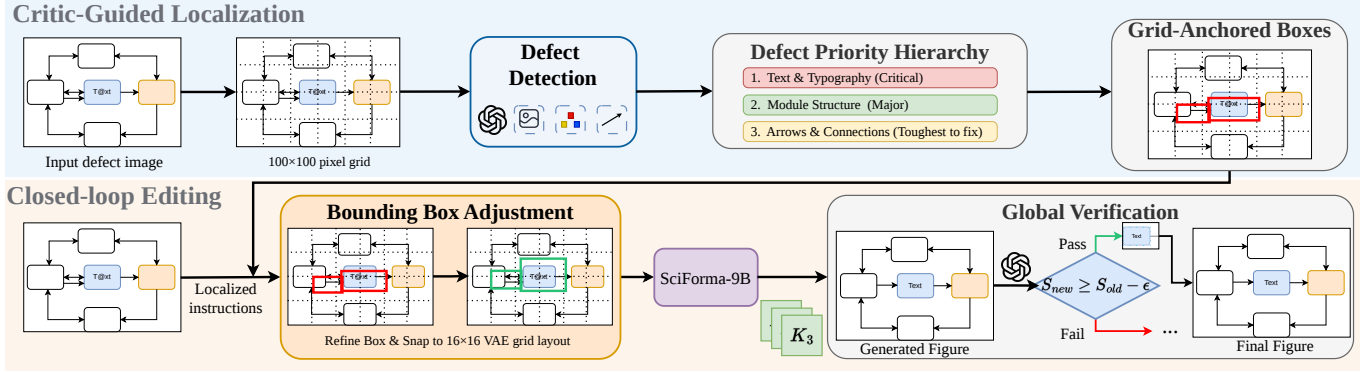}
\Description{Supplemental pipeline figure}
\caption{\textbf{Anatomy of the Closed-Loop Editor.} \textit{Stage 1:} The GPT-5.4 critic uses a $100{\times}100$ visual grid and priority hierarchy to localize defects. The raw bounding box is then refined and snapped to the $16{\times}16$ VAE layout. \textit{Stage 2:} SciForma-9B dynamically generates candidate edits, which are subjected to a global monotonic guard ($\mathcal{S}_{\text{new}} \ge \mathcal{S}_{\text{old}} - \epsilon$). Updates that disrupt holistic layout quality beyond the tolerance margin are strictly rejected.}
\label{fig:supp_refinement}
\end{figure*}

\subsection{Connection to InfoNCE}
\label{app:mdpo:infonce}

The InfoNCE objective for contrastive learning with one positive and $N$ negatives is
\begin{equation}
\mathcal{L}_{\mathrm{InfoNCE}} = -\log\frac{\exp(f(x, x^+))}{\exp(f(x, x^+)) + \sum_{j=1}^{N}\exp(f(x, x^-_j))}.
\end{equation}
Rewriting M-DPO by setting $f(x, y^+) = 0$ (normalizing the winner logit to zero) and $f(x, y^-_d) = -\beta\Delta_d$:
\begin{align}
\mathcal{L}_{\mathrm{M\text{-}DPO}}
&= \log\Bigl(1 + \textstyle\sum_d \exp(-\beta\Delta_d)\Bigr) \nonumber \\
&= -\log\frac{1}{1 + \sum_d \exp(-\beta\Delta_d)} \nonumber \\
&= -\log\frac{\exp(0)}{\exp(0) + \sum_d \exp(-\beta\Delta_d)}.
\end{align}
This is precisely InfoNCE with $N{=}D$ negatives, where each corresponds to an axis-anchored failure mode rather than a randomly sampled distractor. The model succeeds only by generating diagrams distinguishable from every axis-specific failure simultaneously.

Unlike standard InfoNCE where negatives are sampled i.i.d.\ from the data distribution, each M-DPO negative is deliberately mined to isolate a specific structural deficiency, yielding maximally informative gradient directions.

\section{Implementation Details}
\label{app:implementation}

This section provides extended technical details that complement the base architecture and the iterative refinement pipeline introduced in the main text.

\subsection{Base Model Architecture}
\label{app:model_arch}

SciForma builds upon the pre-trained FLUX.2-klein-base-9B~\cite{rename}, utilizing its standard flow-matching Diffusion Transformer (DiT) without introducing structural modifications. Crucially, we leverage its native sequence-concatenation strategy and 4D Rotary Position Embeddings (RoPE) to support joint generation and localized editing. By applying a temporal RoPE offset (where source image tokens receive $T{=}10$ while target tokens retain $T{=}0$), the base model inherently aligns the editing condition with the generation target. This allows us to seamlessly apply our supervised fine-tuning and M-DPO pipelines without needing task-specific heads or architectural bloat.

\subsection{Iterative Refinement Details}
\label{app:iterative_refinement}

To expand on the iterative refinement framework presented in the main text, we map our supplementary technical details directly to its two core stages (visualized in Figure~\ref{fig:supp_refinement}):
\begin{itemize}[leftmargin=*, topsep=2pt, itemsep=1pt]
    \item \textbf{Critic-guided localization:} We detail the defect prioritization hierarchy and introduce the grid-anchored visual protocol for precise bounding box inference.
    \item \textbf{Closed-loop editing:} We describe candidate generation and the global monotonic guard that strictly ensures holistic structural consistency.
\end{itemize}

\noindent\textbf{Critic-guided localization and prioritization.}
The critic uses the GPT-5.4 API to inspect the generated image against the reference caption and structural inventory. The inspection prioritizes defects based on an explicit severity hierarchy to determine a discrete priority score:
\begin{enumerate}[leftmargin=*, topsep=2pt, itemsep=1pt]
    \item \textbf{Text \& Typography (Critical):} Misspelled, entirely missing, or severely garbled text. These are assigned the highest priority as they critically impede readability and are directly actionable via highly focused localized text edits.
    \item \textbf{Module Structure (Major):} Severe rendering distortions, garbled patterns, or broken overlapping elements within structural components.
    \item \textbf{Arrows \& Connections (Minor):} Missing or clearly hallucinated arrows. Minor directional or style variations are strictly tolerated to prevent unnecessary structural disruption.
\end{enumerate}
Based on this inspection, the critic formats localized edit instructions targeting specific sub-regions. The system further triages these targets into either focused \textit{local text edits} or full-region \textit{redraws} depending on defect density and complexity.

\noindent\textbf{Grid-anchored localization protocol.}
Continuous coordinate prediction often truncates core elements. To prevent this, we visually overlay a $100{\times}100$ pixel grid onto the image for the GPT-5.4 critic (see \textbf{Figure~\ref{fig:supp_refinement}}). This anchoring yields highly precise bounding boxes $(x_1, y_1)$ and $(x_2, y_2)$. To ensure VAE compatibility and prevent clipping, we pad and snap these coordinates to the $16{\times}16$ latent grid:
\[
    \tilde{x}_1 = 16 \times \lfloor \max(0, x_1 - p)/16 \rfloor, \quad \tilde{x}_2 = 16 \times \lceil \min(W, x_2 + p)/16 \rceil
\]
where $p$ is a padding margin. We also preserve a $10$-pixel unmasked context strip outside the target box, allowing the model to condition on surrounding local styles.

\noindent\textbf{Closed-loop generation and global guard.}
SciForma-9B dynamically generates up to $K{=}3$ candidate edits per region ($\text{CFG}{=}4.0$). Because localized edits can inadvertently disrupt global connectivity (e.g., an edited box overlapping an adjacent arrow), we implement a whole-image verification guard rather than relying purely on localized evaluations. We define a strict monotonic improvement rule over the holistic quality score $\mathcal{S}$:
\[
    \mathcal{S}_{\text{new}} \ge \mathcal{S}_{\text{old}} - \epsilon
\]
where $\epsilon=0.03$ is a tolerance margin. If a proposed edit causes the overall image quality to drop beyond this tolerance, it is rejected and the system reverts the figure to its previous state (Figure~\ref{fig:supp_refinement}).

\section{SciFormaBench-2K In-depth Analysis}
\label{app:sciformabench_analysis}

In this section, we comprehensively evaluate the robustness, reliability, and validity of our proposed SciFormaBench-2K evaluation protocol. We structure this analysis into three parts: verifying cross-evaluator consistency to address VLM judge bias, confirming multi-round evaluation stability, and demonstrating the orthogonality of our C/A/T scoring axes.

\subsection{Reflexivity Verification}
\label{app:reflexivity}

We first conduct a reflexivity analysis to address the reliability and verifiability of our VLM evaluation paradigm. We score the \textbf{ground-truth reference diagrams} of SciFormaBench-2K against their own human-verified structural inventory. If the evaluator can faithfully recognize the elements it is told to look for, GT diagrams should approach a near-perfect score.
\begin{table}[t!]
\centering
\caption{Reflexivity check: scoring ground-truth diagrams against 
their own human-verified inventory. Near-perfect scores across all 
axes and difficulty tiers confirm the evaluator's extraction reliability.}
\vspace{-3mm}
\label{tab:gt_reflexivity}
\small
\begin{tabular}{lccccccc}
\toprule
Reference & Avg & C & A & T & Simple & Medium & Hard \\
\midrule
GT diagrams & \textbf{99.94} & 99.99 & 99.85 & 99.96 & 99.93 & 99.95 & 99.92 \\
\bottomrule
\end{tabular}
\vspace{-3mm}
\end{table}

Table~\ref{tab:gt_reflexivity} shows GPT-5.4 recovers 99.94\% of checklist elements on average, with per-axis and per-tier scores all above 99.85\%. This confirms the evaluator does not systematically miss present elements, ruling out a common failure mode of VLM-based scoring.

\subsection{Cross-Evaluator Consistency}
\label{app:cross_evaluator}

A valid concern regarding our evaluation protocol is the heavy reliance on a proprietary VLM judge (GPT-5.4) for labeling, evaluation, and verification. This overlap could theoretically introduce evaluator bias, reduce reproducibility, and artificially inflate reported gains. To prove that our structural improvements are fundamental and not merely an artifact of evaluator preference, we fully re-score the SciFormaBench-2K leaderboard using an independent, open-source evaluator: Qwen3-VL-8B-Instruct.

\begin{table}[t!]
\centering
\caption{SciFormaBench-2K re-scored with Qwen3-VL-8B-Instruct. Background: \colorbox{blue!10}{proprietary}, \colorbox{yellow!10}{open-source}, \colorbox{red!10}{ours}.}
\vspace{-3mm}
\label{tab:main_sciformabench_qwen3vl}
\small
\resizebox{\linewidth}{!}{%
\begin{tabular}{lccccccc}
\toprule
Method & Average & Simple & Medium & Hard & C & A & T \\
\midrule
\rowcolor{blue!10} GPT-Image-2               & 70.57 & 80.42 & 70.66 & 65.00 & 67.83 & 72.47 & 74.71 \\
\rowcolor{blue!10} Nano Banana Pro           & 67.09 & 76.56 & 66.89 & 59.50 & 63.16 & 68.73 & 70.53 \\
\rowcolor{red!10} SciForma-9B + Edit       & 61.05 & 70.90 & 60.54 & 54.37 & 57.84 & 62.71 & 63.80 \\
\rowcolor{blue!10} GPT-Image-1.5             & 59.32 & 68.08 & 59.34 & 52.01 & 55.75 & 59.30 & 64.46 \\
\rowcolor{red!10} SciForma-9B              & 59.21 & 69.42 & 58.98 & 51.06 & 56.43 & 60.40 & 61.73 \\
\rowcolor{red!10} SciForma-Base            & 58.65 & 69.24 & 58.16 & 50.54 & 55.30 & 60.36 & 60.60 \\
\rowcolor{yellow!10} Wan2.7-Image                 & 56.75 & 66.83 & 56.66 & 48.46 & 53.03 & 58.19 & 60.09 \\
\rowcolor{yellow!10} Qwen-Image-2512         & 49.76 & 58.62 & 49.71 & 42.45 & 46.86 & 57.27 & 45.36 \\
\rowcolor{yellow!10} Z-Image                 & 49.03 & 57.92 & 48.66 & 42.17 & 42.58 & 55.47 & 49.51 \\
\rowcolor{yellow!10} FLUX.2-dev 32B          & 48.68 & 57.11 & 47.85 & 42.90 & 42.57 & 51.72 & 50.71 \\
\rowcolor{yellow!10} FLUX.2-klein-base-9B    & 38.95 & 43.54 & 37.88 & 36.73 & 36.02 & 41.49 & 37.65 \\
\rowcolor{yellow!10} FLUX.1-dev              & 32.90 & 36.08 & 34.82 & 31.88 & 30.74 & 38.49 & 34.44 \\
\rowcolor{yellow!10} Bagel 7B                & 28.37 & 28.74 & 27.50 & 27.68 & 26.29 & 29.33 & 28.65 \\
\bottomrule
\end{tabular}%
}
\vspace{-5mm}
\end{table}

As shown in Table~\ref{tab:main_sciformabench_qwen3vl}, although absolute scores naturally shift due to differing evaluator strictness, the \textbf{significance of our gains remains completely unaffected}. SciForma-9B + Edit ($61.05$) solidly outperforms all other open-source models—including the far larger FLUX.2-dev 32B ($48.68$) and Qwen-Image-2512 ($49.76$)—and even overtakes proprietary systems like GPT-Image-1.5 ($59.32$). The system-level rankings across both evaluators exhibit near-perfect correlation (Pearson $r{=}0.9945$, Spearman $\rho{=}0.9824$). This definitively confirms that the structural modeling capabilities demonstrated by SciForma represent genuine topological and spatial understanding, completely independent of the chosen VLM judge.

\subsection{Evaluation Robustness across Multiple Rounds}
\label{app:eval_robustness}

To ensure that our automated scoring is not subject to high internal variance, we evaluate the stability of the GPT-5.4 judge across multiple independent runs. Following our standard protocol, we perform the full SciFormaBench-2K evaluation twice (Round 1 and Round 2) for a representative subset of models. 

\begin{table}[t!]
\centering
\caption{Multi-round evaluation stability (GPT-5.4). $|\Delta|$ denotes the absolute score difference between two independent passes.}
\vspace{-3mm}
\label{tab:eval_robustness}
\small
\begin{tabular}{lccc}
\toprule
Model & Round 1 & Round 2 & $|\Delta|$ \\
\midrule
GPT-Image-2 & 70.49 & 70.65 & 0.16 \\
Nano Banana Pro & 67.01 & 67.18 & 0.17 \\
SciForma-9B & 59.18 & 59.24 & 0.06 \\
Wan2.7-Image & 56.68 & 56.82 & 0.14 \\
\bottomrule
\end{tabular}
\vspace{-3mm}
\end{table}

As shown in Table~\ref{tab:eval_robustness}, the absolute difference between the two independent scoring passes is negligible, consistently remaining below $0.22$ points. This empirical evidence confirms that the prompt engineering and structured inventory schema successfully constrain the VLM to produce deterministic, reliable judgments rather than rolling stochastic variations.

\subsection{Independence of C, A, T Evaluation Axes}
\label{app:axis_correlation}

To validate that our three evaluation axes—Component (C), Arrow (A), and Text (T)—capture distinct failure modes, we compute pairwise Pearson correlations ($r$) and shared variance ($R^2$). We evaluate four diverse models spanning commercial APIs (GPT-Image-2), proprietary systems (Nano Banana Pro), open-source diffusion (Wan2.7-Image), and our own intermediate checkpoints, totaling $7{,}391$ valid samples.

\begin{table}[t!]
\centering
\caption{Pairwise Pearson correlation ($r$) and shared variance ($R^2$) between C, A, T axes. Near-zero C--A correlation confirms orthogonal failure modes.}
\vspace{-3mm}
\label{tab:axis_corr}
\small
\setlength{\tabcolsep}{4pt}
\begin{tabular}{l cc cc cc}
\toprule
 & \multicolumn{2}{c}{\textbf{C\,--\,A}} & \multicolumn{2}{c}{\textbf{C\,--\,T}} & \multicolumn{2}{c}{\textbf{A\,--\,T}} \\
\cmidrule(lr){2-3} \cmidrule(lr){4-5} \cmidrule(lr){6-7}
\textbf{Model} & $r$ & $R^2$ & $r$ & $R^2$ & $r$ & $R^2$ \\
\midrule
GPT-Image-2 & $0.013$ & 0.0\% & $0.080^{***}$ & 0.6\% & $0.091^{***}$ & 0.8\% \\
Nano Banana Pro & $0.004$ & 0.0\% & $0.151^{***}$ & 2.3\% & $0.181^{***}$ & 3.3\% \\
SciForma-Base & $0.009$ & 0.0\% & $0.140^{***}$ & 2.0\% & $0.181^{***}$ & 3.3\% \\
Wan2.7-Image & $0.015$ & 0.0\% & $0.209^{***}$ & 4.4\% & $0.264^{***}$ & 7.0\% \\
\bottomrule
\end{tabular}
\vspace{1mm}
{\footnotesize $^{***}\!p{<}0.001$;\; all unmarked pairs have $p{>}0.05$. $R^2$ denotes shared variance (\%).}
\vspace{-4mm}
\end{table}

Table~\ref{tab:axis_corr} reveals that the Component and Arrow axes are effectively orthogonal ($r < 0.02$, $p > 0.05$) across all models. The remaining pairs (C--T, A--T) show only weak positive correlations ($r \le 0.26$) with maximum shared variance strictly below $7\%$. This stable pattern confirms the evaluation rubric isolates distinct structural properties independent of specific model capabilities.

\textbf{Crucially, this structural orthogonality heavily impacts online RL reward design.} If these three axes are statistically independent, a naive averaged scalar reward $\bar{r} = (r_C + r_A + r_T)/3$ severely dilutes the learning gradient. For example, a model that systematically misplaces arrows but generates correct components receives an aggregated reward dominated by the C and T terms. This leaves negligible learning pressure to fix the critical arrow deficiency. This empirical mismatch strongly motivates our multi-dimensional conjunctive reward formulation (Section~\ref{app:mdpo_analysis}), which actively prevents reward hacking across orthogonal dimensions by enforcing strict per-axis bottlenecks.

\section{More Ablations}
\label{app:more_ablations}

This section introduces additional ablations that isolate specific modeling choices. We first ablate SFT Stage 2 editing data to observe its effect on multi-round iterative refinement (Section~\ref{app:sft_ablations}). Subsequently, we detail M-DPO ablations (Section~\ref{app:mdpo_ablations}), which explore preference win-lose pair curation mechanics, the multi-dimensional decomposition of contrastive axes, and hyperparameters such as the KL divergence penalty $\beta$.

\subsection{SFT Ablations}
\label{app:sft_ablations}

\subsubsection{Editing Triplet Supervision (Stage~2)}
\label{app:editing_triplets}

SciFormaData-700K contains 70K editing triplets alongside generation pairs. Since these triplets target localized edits rather than full-image generation, their benefit only manifests through the downstream iterative editing loop. We train two SFT checkpoints to 60K steps—one with and one without editing triplets—then apply the same iterative editing budget to both.

\begin{table}[t!]
\centering
\caption{Effect of editing triplet supervision on iterative refinement quality (SciFormaBench-2K, post-edit).}
\label{tab:ablation_edit_triplets}
\small
\begin{tabular}{lcccc}
\toprule
Setup & Avg & C & A & T \\
\midrule
w/o edit triplets + editing & 68.76 & 73.62 & 66.17 & 65.81 \\
w/ edit triplets + editing & \textbf{71.72} & \textbf{75.30} & \textbf{68.61} & \textbf{68.47} \\
\midrule
$\Delta$ & \textbf{+2.96} & +1.68 & +2.44 & +2.66 \\
\bottomrule
\end{tabular}
\end{table}

Including editing triplets during SFT yields a $+2.96$ average improvement over the ablated model after the same editing budget. The gains are consistent across all three axes, with Arrow ($+2.44$) and Text ($+2.66$) benefiting most. This confirms that editing supervision is essential: without it, the model lacks the localized repair capability needed for effective inference-time refinement.

\subsection{M-DPO Ablations}
\label{app:mdpo_ablations}

\subsubsection{Preference Pair Construction Strategy}
\label{app:preference_construction}

Effective preference optimization requires high-quality, diverse preference pairs.
Before examining the axis-specific loss design, we first ablate the \textit{data construction strategy} that underlies all M-DPO training.
Table~\ref{tab:dpo_full_comparison} compares four strategies, varying both the winner source and the rollout diversity, applied to the same SFT checkpoint.

\begin{table}[t!]
\centering
\caption{Preference pair construction strategy ablation on SciFormaBench-2K (GPT-5.4). We compare winner sources and rollout diversity settings.}
\label{tab:dpo_full_comparison}
\small
\resizebox{\linewidth}{!}{
\begin{tabular}{lcccc}
\toprule
Model / Config & Average & C & A & T \\
\midrule
SciForma-Base & 67.59 & 73.52 & 64.64 & 63.84 \\
\midrule

\shortstack[l]{+ DPO\\(GT win)} & 66.57\textcolor{red!80!black}{(-1.02)} & 72.74\textcolor{red!80!black}{(-0.78)} & 63.09\textcolor{red!80!black}{(-1.55)} & 63.14\textcolor{red!80!black}{(-0.70)} \\

\shortstack[l]{+ DPO\\(GPT1.5 win)} & 67.11\textcolor{red!80!black}{(-0.48)}  & 73.43\textcolor{red!80!black}{(-0.09)}  & 63.28\textcolor{red!80!black}{(-1.36)}  & 63.98\textcolor{green!60!black}{(+0.14)}  \\

\shortstack[l]{+ DPO\\(4$\times$25 step)} & 67.71\textcolor{green!60!black}{(+0.12)} & 73.48\textcolor{red!80!black}{(-0.04)} & 64.01\textcolor{red!80!black}{(-0.63)} & 65.03\textcolor{green!60!black}{(+1.19)} \\

\shortstack[l]{+ DPO\\(4$\times$50+8$\times$25 step)} &
\textbf{68.42}\textcolor{green!60!black}{(+0.83)} & \textbf{74.14}\textcolor{green!60!black}{(+0.62)} & \textbf{65.32}\textcolor{green!60!black}{(+0.68)} & \textbf{65.08}\textcolor{green!60!black}{(+1.24)} \\
\bottomrule
\end{tabular}
}
\end{table}

The results expose a decisive hierarchy among construction strategies.
\textbf{(i)}~Using ground-truth images as winners (GT~win) \textit{degrades} all three axes ($-1.02$ average).
We attribute this to the distribution gap between model-generated TikZ and human-authored reference code: the model cannot extract a learnable preference signal from out-of-distribution winners.
\textbf{(ii)}~Replacing the winner source with GPT-selected best-of-$N$ candidates (GPT1.5~win) narrows the gap yet still yields a net loss ($-0.48$), with the Component and Arrow axes remaining below the SFT baseline.
\textbf{(iii)}~Uniform short rollouts ($4{\times}25$ step) produce near-baseline results ($+0.12$ average).
Although the Text axis receives a noticeable boost ($+1.19$), the limited generation diversity fails to surface meaningful contrastive signals for Component and Arrow.
\textbf{(iv)}~Only the \textit{long-short mixing} strategy ($4{\times}50{+}8{\times}25$ step) achieves substantial, balanced gains across all axes ($+0.83$ average), with improvements of $+0.62$, $+0.68$, and $+1.24$ on C, A, and T respectively.
By combining long rollouts that expose coarse layout and structural errors with short rollouts that reveal fine-grained detail and arrow-placement mistakes, this strategy produces preference pairs spanning the full structural spectrum.

\textit{The long-short rollout strategy provides the essential training data for all subsequent M-DPO experiments} (Table~\ref{tab:ablation_num_axes_expanded}).
Without it, the contrastive training signal is either degenerate (rows~1--2) or too narrow (row~3) for preference optimization to improve over the SFT baseline.

\subsubsection{Number of Contrastive Axes ($D$)}
\label{app:num_axes}

Given the long-short data pipeline established above, we now isolate the contribution of the axis-specific loss by varying the number of contrastive axes~$D$.
At $D{=}1$, M-DPO reduces to standard scalar DPO (Appendix~\ref{app:mdpo:reduction}), optimizing against a single axis-specific loser.
We exhaustively enumerate all axis subsets to reveal how each axis contributes individually and in combination.

\begin{table}[t!]
\centering
\caption{Detailed ablation of contrastive axes $D$ on SciFormaBench-2K (GPT-5.4). We compare single-axis ($D=1$), dual-axis ($D=2$), and triple-axis ($D=3$) configurations. The intensity of the green color indicates the relative magnitude of improvement within each column.}
\label{tab:ablation_num_axes_expanded}
\small
\definecolor{MyGreen}{rgb}{0.0, 0.5, 0.0} 
\resizebox{\linewidth}{!}{
\begin{tabular}{lcccc}
\toprule
Method & Average & C & A & T \\
\midrule
SciForma-Base & 67.59 & 73.52 & 64.64 & 63.84 \\
\midrule
$D=1$ (C) & 68.14\textcolor{MyGreen!55}{(+0.55)} & 74.29\textcolor{MyGreen!65}{(+0.77)} & 65.32\textcolor{MyGreen!55}{(+0.68)} & 64.21\textcolor{MyGreen!55}{(+0.37)} \\
$D=1$ (A) & 68.19\textcolor{MyGreen!57}{(+0.60)} & 74.25\textcolor{MyGreen!55}{(+0.73)} & 65.85\textcolor{MyGreen!75}{(+1.21)} & 64.19\textcolor{MyGreen!55}{(+0.35)} \\
$D=1$ (T) & 67.21\textcolor{red!80!black}{(-0.38)} & 73.12\textcolor{red!80!black}{(-0.40)} & 64.01\textcolor{red!80!black}{(-0.63)} & 64.41\textcolor{MyGreen!60}{(+0.57)} \\
\midrule

$D=2$ (C, A) & 68.38\textcolor{MyGreen!66}{(+0.79)} & 74.43\textcolor{MyGreen!84}{(+0.91)} & 65.53\textcolor{MyGreen!67}{(+0.89)} & 64.50\textcolor{MyGreen!62}{(+0.66)} \\
$D=2$ (C, T) & 68.90\textcolor{MyGreen!81}{(+1.31)} & 74.42\textcolor{MyGreen!81}{(+0.90)} & 65.46\textcolor{MyGreen!63}{(+0.82)} & 65.54\textcolor{MyGreen!76}{(+1.70)} \\
$D=2$ (A, T) & 68.49\textcolor{MyGreen!72}{(+0.90)} & 74.36\textcolor{MyGreen!75}{(+0.84)} & 66.14\textcolor{MyGreen!82}{(+1.50)} & 64.88\textcolor{MyGreen!71}{(+1.04)} \\
\midrule
\shortstack[l]{$D=3$ (C, A, T) \\ (SciForma-9B)} & \textbf{69.51}\textcolor{MyGreen!100}{(+1.92)} & \textbf{74.49}\textcolor{MyGreen!100}{(+0.97)} & \textbf{66.46}\textcolor{MyGreen!100}{(+1.82)} & \textbf{67.00}\textcolor{MyGreen!100}{(+3.16)} \\
\bottomrule
\end{tabular}
}
\end{table}

Several clear patterns emerge from the full enumeration.
\textbf{(i)}~Single-axis optimization ($D{=}1$) tends to improve the targeted dimension but provides limited, or even negative, transfer to the other two, confirming that structural correctness~(C), spatial accuracy~(A), and textual fidelity~(T) demand distinct contrastive signals that a single scalar objective cannot simultaneously capture.
\textbf{(ii)}~Dual-axis configurations ($D{=}2$) consistently outperform their corresponding single-axis counterparts on the Average metric, yet the un-contrasted third axis lags behind, demonstrating that two axes alone are insufficient for conjunctive correctness.
\textbf{(iii)}~Performance improves monotonically as $D$ increases: $D{=}1 < D{=}2 < D{=}3$, with the full triple-axis configuration (C,\,A,\,T) achieving the highest average score. Gains are distributed across all three axes, confirming the theoretical prediction of Section~\ref{app:mdpo_analysis}: the conjunctive loss automatically concentrates the gradient on whichever axis is weakest in each training batch, yielding balanced improvement without manual axis weighting.

\subsubsection{KL Divergence Penalty $\beta$}
\label{app:beta_sensitivity}

The KL divergence penalty $\beta$ controls the strength of preference optimization: larger $\beta$ pushes the policy more aggressively toward preferred samples, while smaller $\beta$ keeps updates closer to the reference SFT model. We sweep $\beta$ across a $3\times$ range to 
characterize sensitivity.

\begin{table}[t!]
\centering
\caption{$\beta$ sensitivity ablation on SciFormaBench-2K (GPT-5.4).}
\vspace{-3mm}
\label{tab:beta_ablation}
\small
\begin{tabular}{lcccc}
\toprule
$\beta$ & Avg & C & A & T \\
\midrule
$1000$ & $68.73$ & $\textbf{75.01}$ & $65.92$ & $64.46$ \\
$2000$ (used) & $\textbf{69.51}$ & $74.49$ & $\textbf{66.46}$ & $\textbf{67.00}$ \\
$3000$ & $68.58$ & $74.31$ & $64.75$ & $65.73$ \\
\bottomrule
\end{tabular}
\vspace{-3mm}
\end{table}

Table~\ref{tab:beta_ablation} reveals two patterns: \textbf{(i)}~M-DPO is robust across an order of magnitude in $\beta$, with all three settings clearing the SFT baseline. \textbf{(ii)}~The Text axis is most sensitive—too-small $\beta$ under-penalizes weak-axis errors, while too-large $\beta$ saturates the gate. Our default $\beta{=}2000$ sits at the optimal balance, delivering the highest gains on the weakest axes ($\text{A}\,{+}1.82$, $\text{T}\,{+}3.16$).

\section{Comparison with Code-Based Baselines}
\label{app:code_based}

We compare SciForma against two representative code-based diagram-generation approaches on SciFormaBench-2K: \textbf{TikZilla}~\cite{TikZilla}, an LLM-based TikZ code synthesis pipeline, and \textbf{GLM-4.7-Flash}, a general-purpose LLM prompted to produce LaTeX/TikZ diagram code.

\begin{table}[t!]
\centering
\caption{Code-based baselines on SciFormaBench-2K. Both score 
substantially below SciForma-9B, limited by template-driven renderer 
constraints.}
\vspace{-3mm}
\label{tab:code_baselines}
\small
\begin{tabular}{lcccc}
\toprule
Method & Avg & C & A & T \\
\midrule
TikZilla        & $35.81$ & $41.08$ & $30.50$ & $34.99$ \\
GLM-4.7-Flash   & $38.21$ & $46.93$ & $32.61$ & $33.87$ \\
\midrule
SciForma-9B (ours) & $\textbf{69.51}$ & $\textbf{74.49}$ & $\textbf{66.46}$ & $\textbf{67.00}$ \\
\bottomrule
\end{tabular}
\vspace{-3mm}
\end{table}

Table~\ref{tab:code_baselines} shows that both code-based methods score roughly \textit{half} of SciForma ($35.81\,/\,38.21$ vs.\ $69.51$~Avg). Manual inspection reveals three dominant failure modes: \textbf{(i)}~layout overlap and cluttering on complex diagrams, \textbf{(ii)}~systematic arrow endpoint mismatch, and \textbf{(iii)}~outright rendering failures. This confirms that structural fidelity in free-form scientific diagrams requires pixel-space generation with structural supervision, rather than syntactic code synthesis.

\section{Additional Qualitative Results}
\label{app:qualitative_results}

This section provides extended visual comparisons of SciForma-9B against both proprietary and open-source baselines, followed by iterative refinement examples and failure case analysis.

\subsection{Generation Gallery}
\label{app:generation_gallery}

\subsubsection{Comparison with Proprietary Models}
\label{app:qualitative_proprietary}

\begin{figure*}[t!]
    \centering
    \Description{Qualitative comparison with proprietary models (1/2).}
    \includegraphics[width=0.98\linewidth]{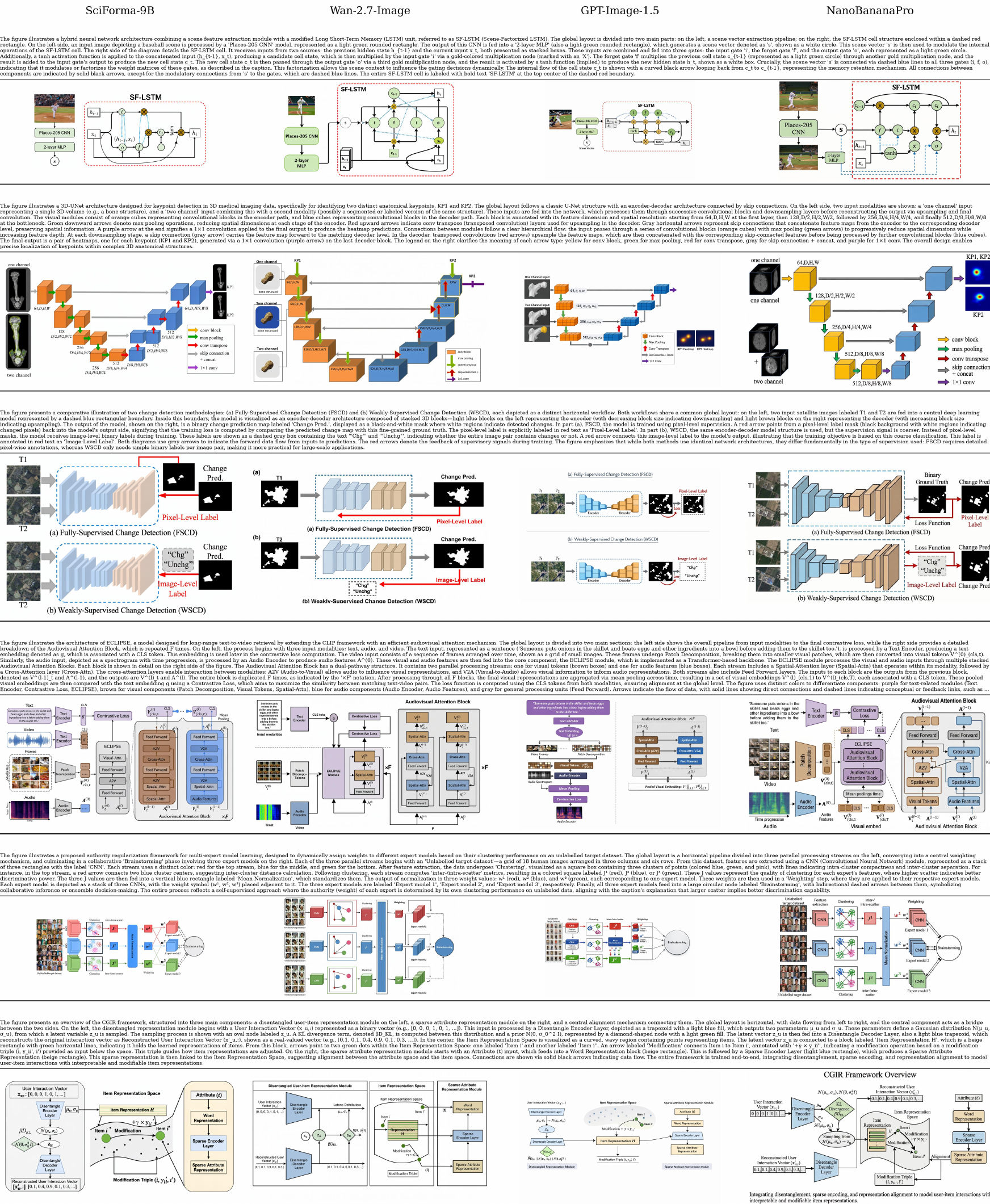}
    \vspace{-3mm}
    \caption{Qualitative comparison with proprietary models (1/2). SciForma-9B vs.\ Wan2.7-Image, GPT-Image-1.5, and Nano Banana Pro on the same prompts.}
    \label{fig:proprietary_page1}
\end{figure*}

\begin{figure*}[t!]
    \centering
    \Description{Qualitative comparison with proprietary models (2/2).}
    \includegraphics[width=0.98\linewidth]{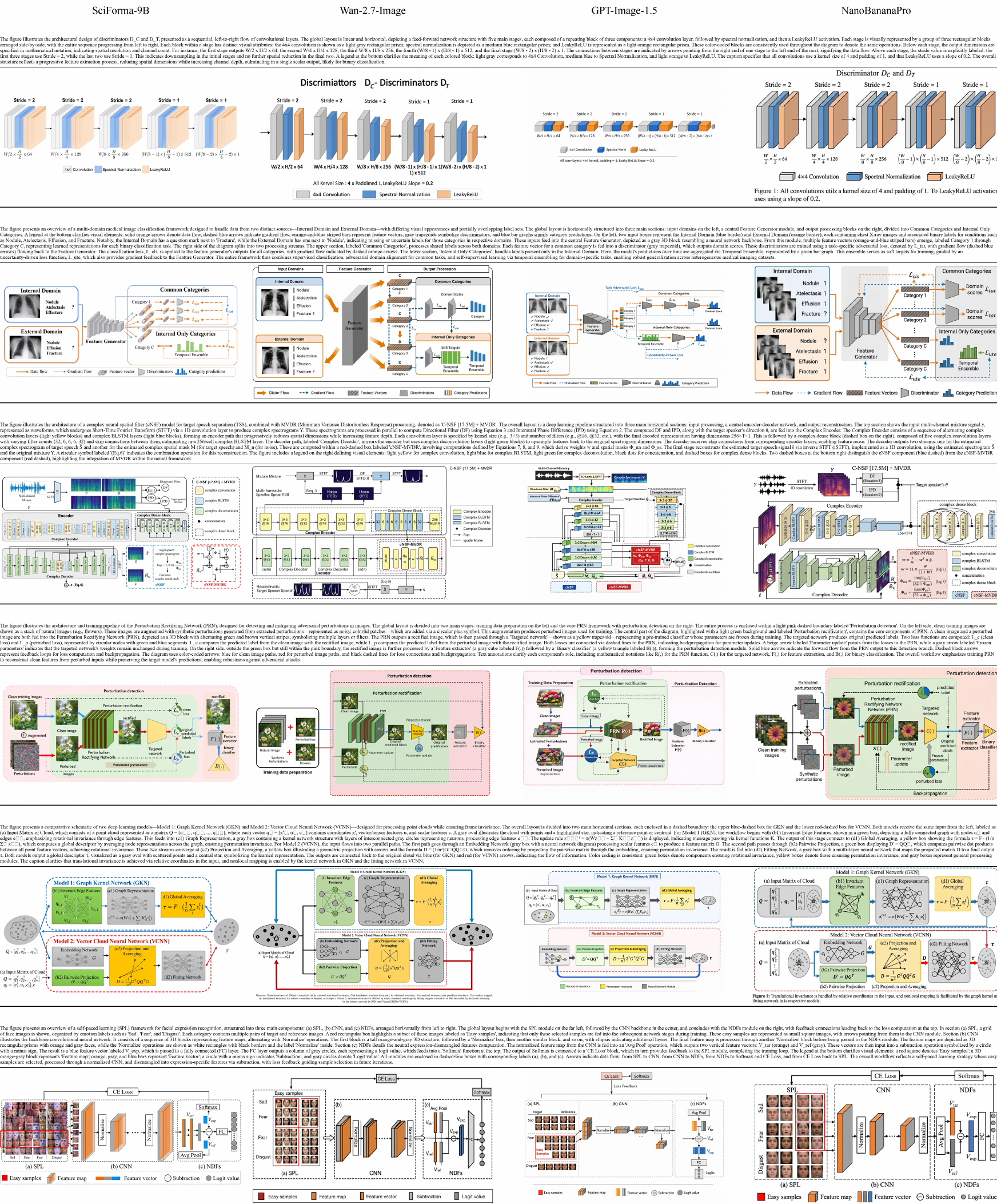}
    \vspace{-3mm}
    \caption{Qualitative comparison with proprietary models (2/2). Same setup as Figure~\ref{fig:proprietary_page1}.}
    \label{fig:proprietary_page2}
\end{figure*}

Figures~\ref{fig:proprietary_page1} and~\ref{fig:proprietary_page2} juxtapose SciForma-9B outputs with three proprietary systems---Wan2.7-Image, GPT-Image-1.5, and Nano Banana Pro---on architecture diagrams sampled from SciFormaBench-2K.

Three systematic differences stand out.
\textbf{(i)~Aspect-ratio flexibility.}
SciForma-9B produces images at arbitrary aspect ratios that match the natural proportions of the described diagram (e.g., wide panoramic pipelines, tall encoder--decoder stacks).
In contrast, GPT-Image-1.5 is restricted to three fixed canvas sizes ($1024{\times}1024$, $1024{\times}1536$, $1536{\times}1024$), frequently leaving blank margins when the content does not fill the chosen canvas.
\textbf{(ii)~Visual style.}
Because SciForma is trained on real arXiv figures, its color palettes, line weights, and typographic conventions closely resemble human-authored diagrams.
Wan2.7-Image and Nano Banana Pro, by contrast, tend to render heavy outlines around every module, producing a stylized, poster-like appearance that departs noticeably from academic publishing norms.
\textbf{(iii)~Layout fidelity.}
SciForma preserves fine-grained spatial relationships---skip connections, residual arrows, multi-branch splits---more faithfully than all three competitors, consistent with its quantitative advantage on the Arrow axis reported in the main text.

\subsubsection{Comparison with Open-Source Models}
\label{app:qualitative_opensource}

\begin{figure*}[t!]
    \centering
    \Description{Qualitative comparison with open-source models (1/2).}
    \includegraphics[width=0.98\linewidth]{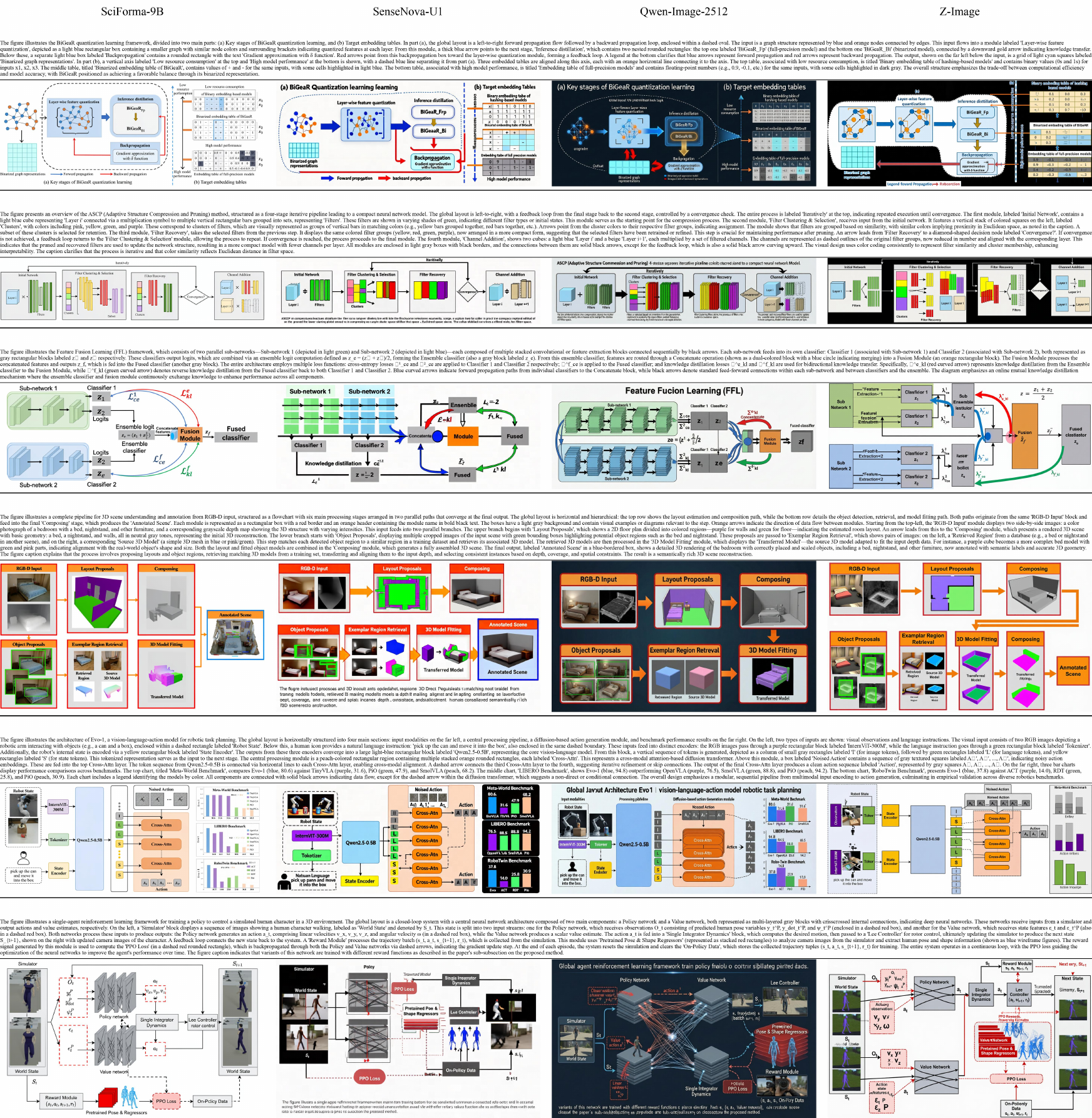}
    \vspace{-3mm}
    \caption{Qualitative comparison with open-source models (1/2). SciForma-9B vs.\ SenseNova-U1, Qwen-Image-2512, and Z-Image.}
    \label{fig:opensource_page1}
\end{figure*}

\begin{figure*}[t!]
    \centering
    \Description{Qualitative comparison with open-source models (2/2).}
    \includegraphics[width=0.98\linewidth]{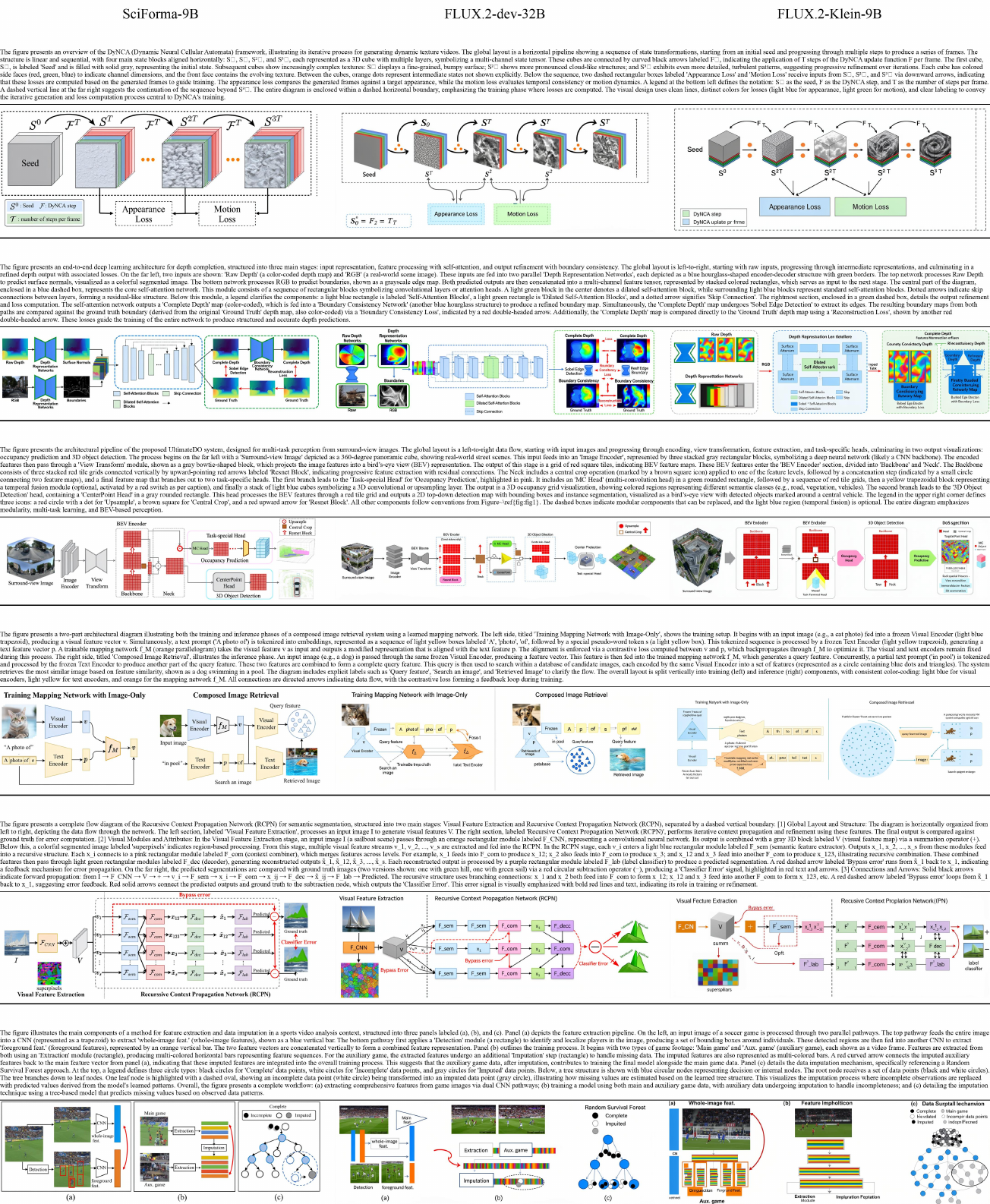}
    \vspace{-3mm}
    \caption{Qualitative comparison with open-source models (2/2). SciForma-9B vs.\ FLUX.2-dev-32B and FLUX.2 Klein Base-9B.}
    \label{fig:opensource_page2}
\end{figure*}

Figures~\ref{fig:opensource_page1} and~\ref{fig:opensource_page2} compare SciForma-9B against open-source baselines spanning unified multimodal models (SenseNova-U1, Qwen-Image-2512, Z-Image) and the FLUX family (FLUX.2-dev-32B, FLUX.2 Klein Base-9B).

\textbf{Qwen-Image-2512 and Z-Image} struggle with basic structural elements: shapes are malformed, arrows are frequently misplaced or missing, and outputs often exhibit dark or black backgrounds far removed from the visual distribution of scientific figures.
\textbf{FLUX.2-dev-32B and FLUX.2 Klein Base-9B} produce more aesthetically pleasing color schemes---indeed, this is a key reason we chose the FLUX architecture as our base model. However, FLUX renders text poorly and its shapes lack structural common sense. Crucially, while FLUX outputs appear plausible at first glance, closer inspection reveals pervasive detail errors: misconnected arrows, phantom modules, and incorrect spatial hierarchies. This explains why FLUX.2-dev-32B, despite being 32B parameters, scores comparably to Qwen-Image-2512 and Z-Image on SciFormaBench-2K.
\textbf{SenseNova-U1}, a recent unified multimodal model, demonstrates strong structural understanding and is the most competitive open-source baseline. Its main weakness is aesthetic: outputs tend toward flat coloring and lack the polished appearance of human-authored diagrams. With more careful data curation, this architecture family could become a strong contender for scientific figure generation.

\subsection{Iterative Refinement and Alignment}
\label{app:qualitative_refinement_alignment}

\begin{figure*}[t!]
    \centering
    \Description{Qualitative illustration of editing.}
    \includegraphics[width=\linewidth]{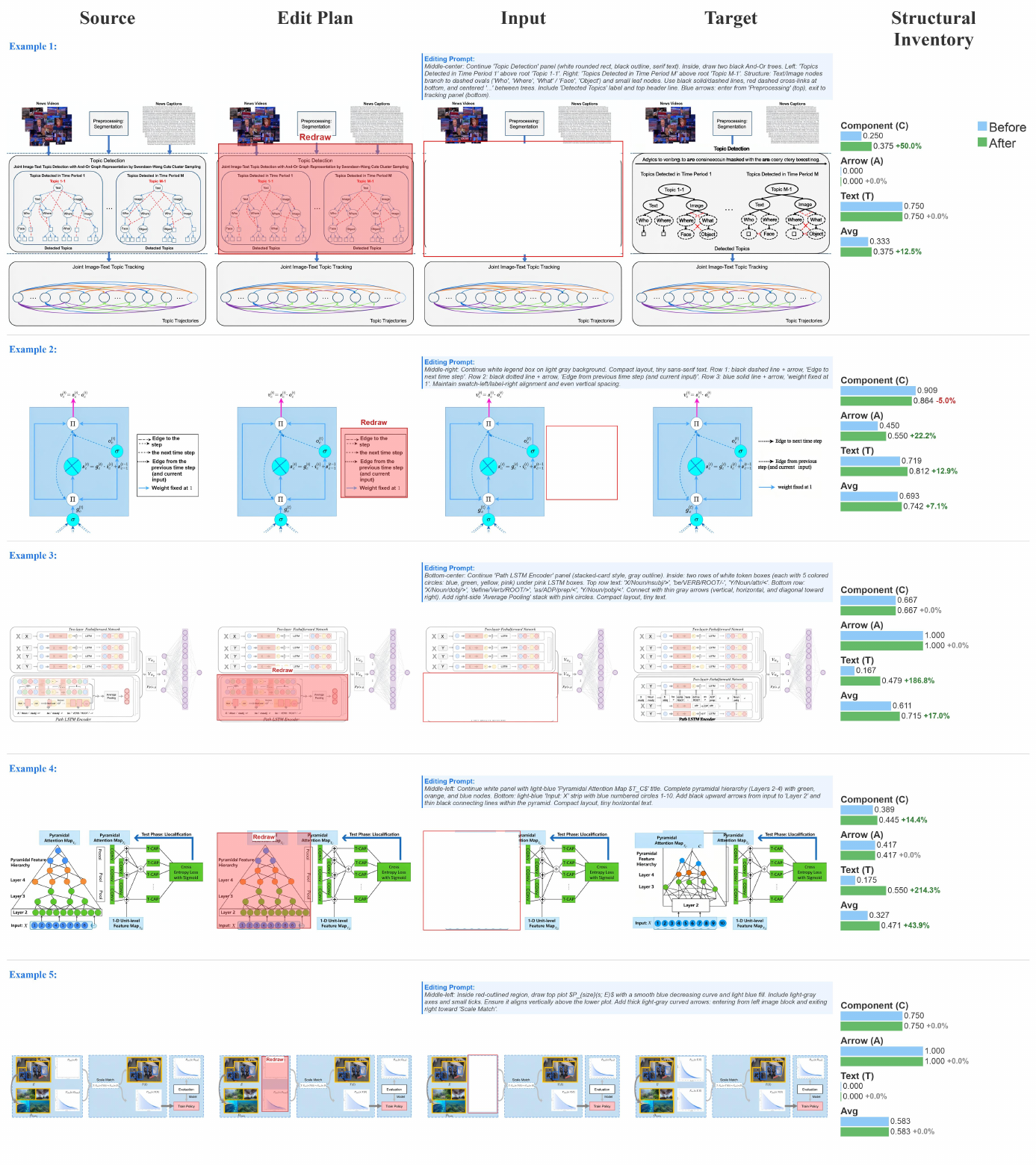}
    \vspace{-3mm}
    \caption{Further examples of iterative visual refinement. SciForma-9B accurately updates localized structures, texts, and arrow connections based on editing instructions.}
    \label{fig:sppl_editing_examples}
\end{figure*}

Figure~\ref{fig:sppl_editing_examples} provides five additional examples of the iterative refinement process. Guided by localized edit instructions generated by GPT-5.4, SciForma-9B accurately refines the designated region and redraws it to incorporate clear structural components, direct arrows, and accurate texts. This highlights the model's robust spatial control and iterative refinement capabilities.

\begin{figure*}[t!]
    \centering
    \Description{M-DPO qualitative results}
    \includegraphics[width=\linewidth]{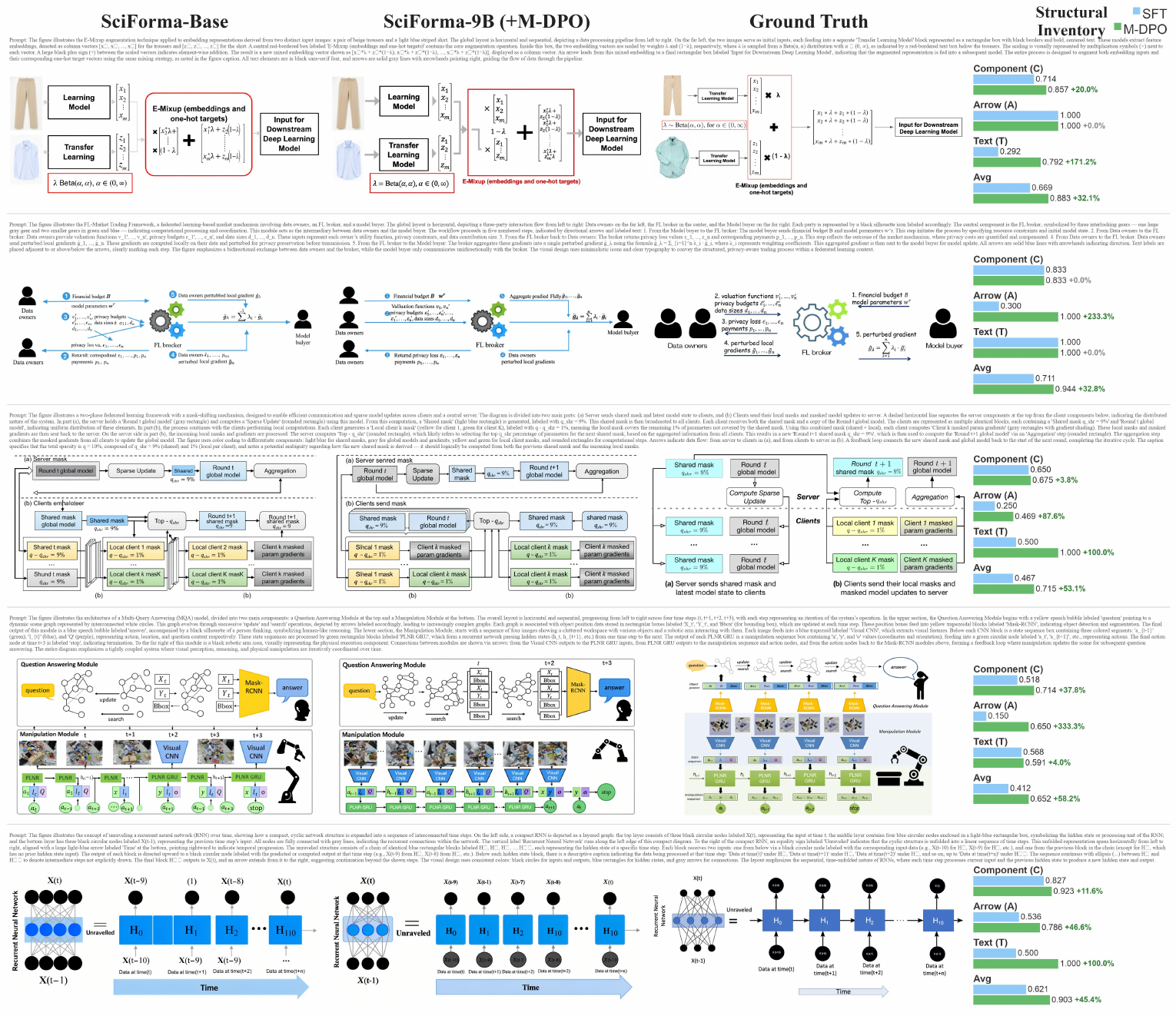}
    \vspace{-3mm}
    \caption{Qualitative effects of M-DPO. The preference-aligned model (SciForma-9B + M-DPO) effectively corrects structural errors, repairs arrow connections, and improves text placement compared to the base model.}
    \label{fig:sppl_mdpo_effects}
\end{figure*}

Figure~\ref{fig:sppl_mdpo_effects} provides visual comparisons demonstrating the improvements brought by M-DPO. Compared to the supervised fine-tuning baseline (SciForma-Base), the M-DPO-aligned model reduces structural hallucinations, repairs broken or misaligned arrow connections, and refines text layouts, thereby producing diagrams that more faithfully align with the ground truth.

\subsection{Failure Cases}
\label{app:failure_cases}

\begin{figure}[t!]
    \centering
    \Description{Failure cases of SciForma-9B organized into three categories: rotated text corruption, arrow errors in complex topologies, and bounding box bottlenecks in iterative editing.}
    \includegraphics[width=0.99\linewidth]{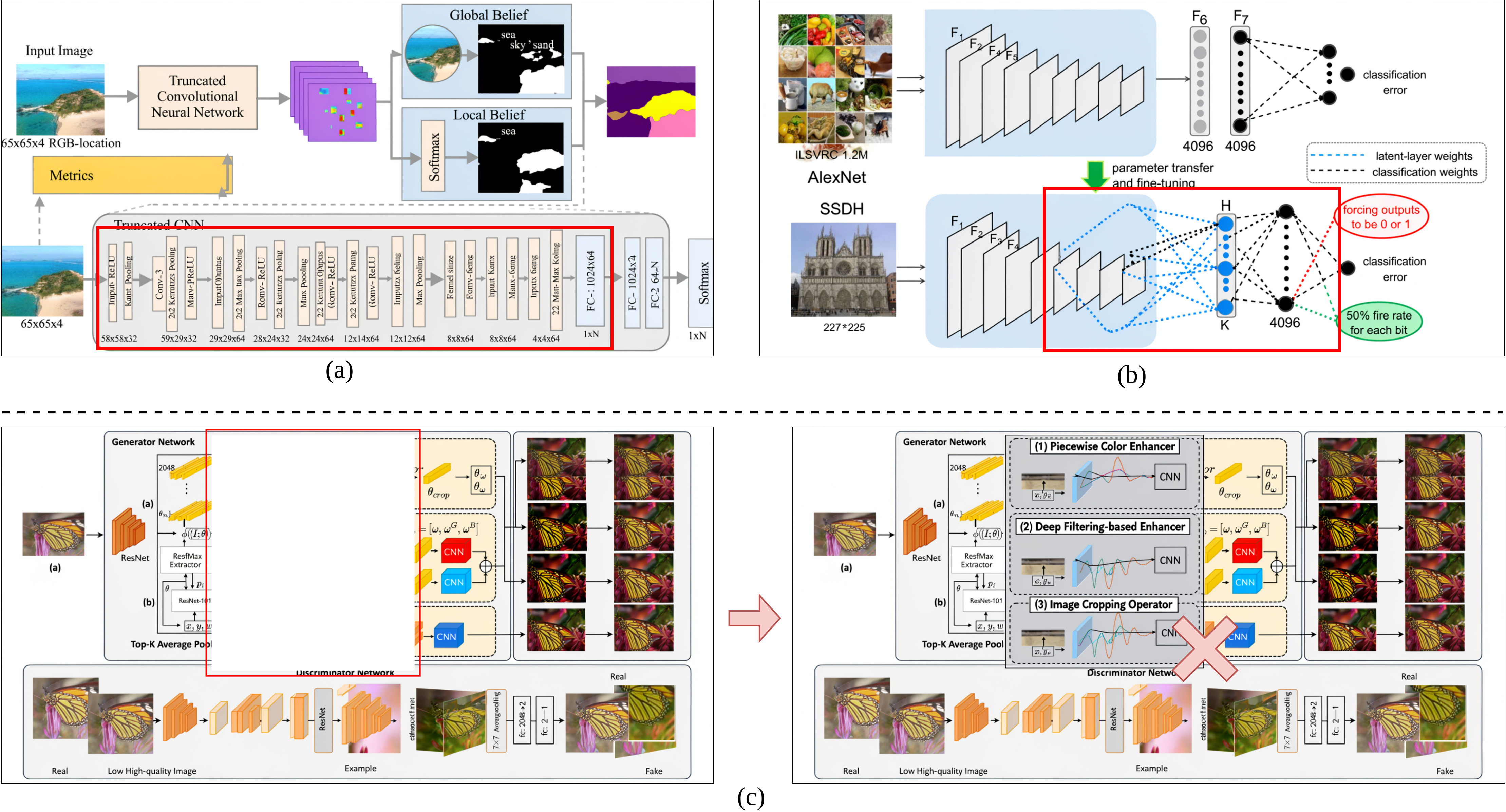}
    \vspace{-3mm}
    \caption{Representative failure cases of SciForma-9B. (a)~Rotated text: vertically oriented labels are garbled or collapsed to horizontal. (b)~Arrow errors in complex topologies: spurious, missing, or misrouted connections in dense diagrams. (c)~Iterative editing bottlenecks: imprecise bounding box detection and cropping in dense regions lead to uncoordinated inpainting and introduce new visual artifacts.}
    \label{fig:failure_cases}
\end{figure}

Despite strong overall fidelity, SciForma-9B exhibits three persistent failure modes (Figure~\ref{fig:failure_cases}):

\textbf{(i) Text rendering failures.} It struggles with non-horizontal text (e.g., rotated y-axis labels), often garbling characters or defaulting to horizontal orientations due to base-model biases.

\textbf{(ii) Arrow errors in dense topologies.} Highly complex or overlapping connections occasionally cause spurious, missing, or misattached arrows, especially under underspecified routing prompts.

\textbf{(iii) Bounding-box bottlenecks in iterative editing.} In dense diagrams, imprecise bounding-box cropping often truncates adjacent elements. This leads to uncoordinated inpainting that clashes with the global context, introducing new artifacts.

Future work includes incorporating rotated-text data, adding graph-level topology constraints to reduce arrow ambiguity, and developing context-aware region selection (e.g., segmentation-driven editing) to bypass rigid bounding boxes.

\subsection{User Study Details}
\label{app:user_study_details}

To validate that structural-inventory metrics reflect perceptible quality differences, we conduct a pairwise preference study. We sample 50 prompts from SciFormaBench-2K, stratified by difficulty (17~Simple, 17~Medium, 16~Hard), and render each with three systems: SciForma-9B, Z-Image (strongest open-source baseline), and GPT-Image-1.5 (strongest proprietary baseline below GPT-Image-2). For each comparison pair, annotators view the reference diagram alongside two anonymized, randomly ordered generated images and select a winner or draw on four perceptual axes:
{\emergencystretch=1em
\begin{itemize}[leftmargin=*, topsep=2pt, itemsep=2pt, parsep=0pt]
    \item \textbf{Faithfulness}: \emph{Does the generated diagram convey the same methodology as the reference?} Annotators assess whether the key concepts, modules, and their logical relationships are correctly represented, judging holistic semantic fidelity rather than pixel-level similarity.
    \item \textbf{Readability}: \emph{Can you follow the diagram without effort?} This covers text legibility, unambiguous flow direction, and whether the layout naturally guides the eye through the pipeline.
    \item \textbf{Aesthetics}: \emph{Is the diagram publication-ready?} Annotators evaluate visual polish: clean shapes, coherent color scheme, consistent styling, and absence of rendering artifacts.
    \item \textbf{Overall}: \emph{Which diagram would you include in your paper?}
\end{itemize}
\par}

\noindent Each pair is evaluated by three annotators with graduate-level research experience in CS/AI; the majority vote determines the label. We report Fleiss' $\kappa$ to quantify inter-annotator agreement.

\begin{figure}[t!]
    \centering
    \Description{User study questionnaire interface.}
    \includegraphics[width=\linewidth]{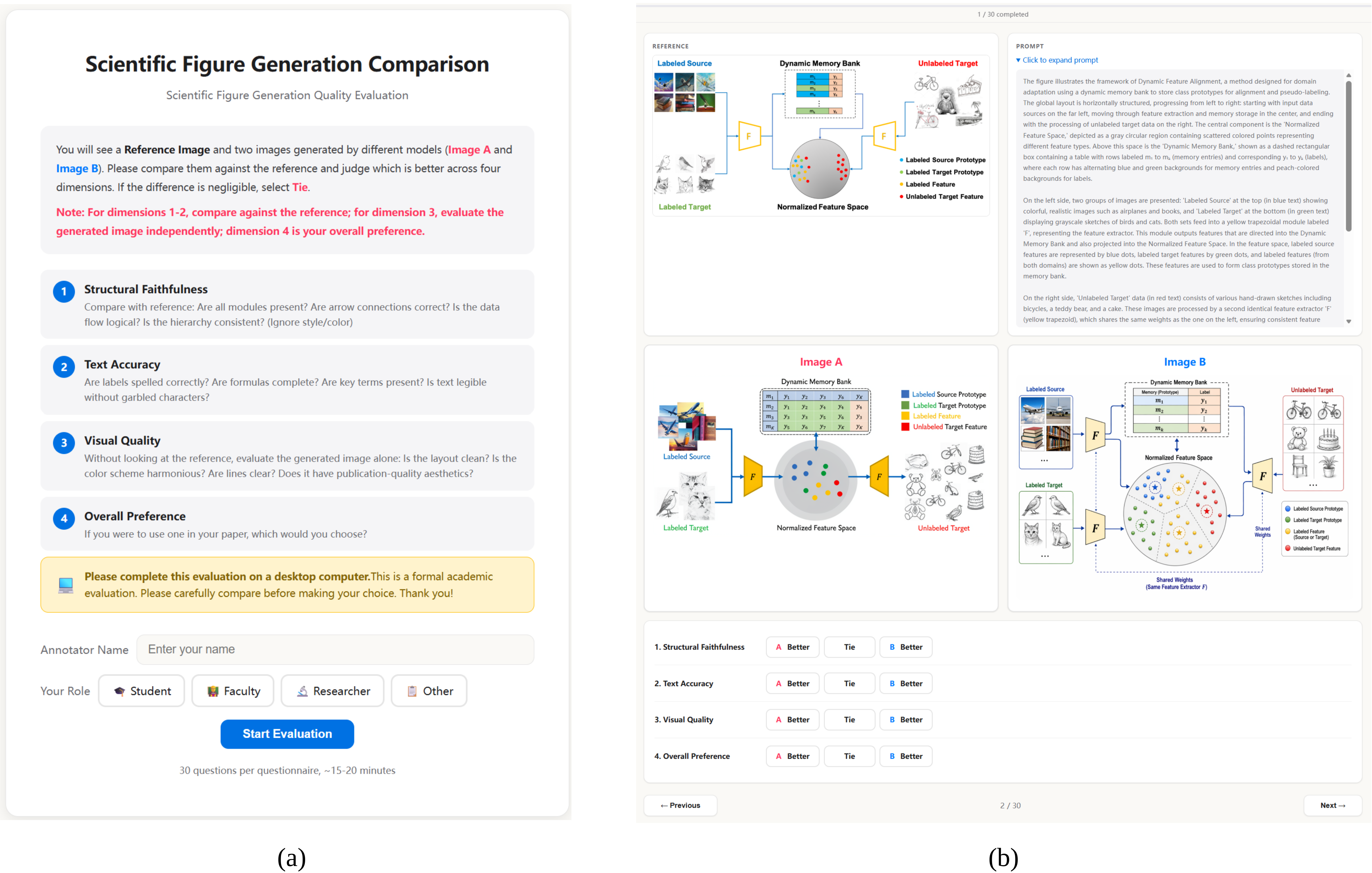}
    \vspace{-3mm}
    \caption{User study questionnaire screenshot. (a) Study instructions outlining the evaluation dimensions. (b) The pairwise comparison interface, displaying the prompt, the reference diagram, and two anonymized generated diagrams (Image A and Image B).}
    \label{fig:sppl_user_study_interface}
    \Description{sppl_user_study_interface}
\end{figure}

\clearpage
\section{Prompt Templates}
\label{app:all_prompts}

This section provides the exact prompt templates used for data construction (Section~\ref{app:data_prompts}) and structural evaluation (Section~\ref{app:eval_prompt_templates}), enabling full reproducibility of the SciForma pipeline.

\subsection{Data Pipeline Prompts}
\label{app:data_prompts}

\noindent\textbf{LaTeX macro resolver (Stage~1).}
\begin{tcolorbox}[title={LaTeX Macro Resolver --- System Prompt}, fonttitle={\bfseries\footnotesize}, fontupper=\scriptsize, colback={gray!5}, colframe={gray!60}, coltitle={white}]
\texttt{You are an advanced LaTeX Code Analyzer. Your task is to parse raw LaTeX `figure` environments to extract structured metadata, specifically resolving local macros.}

\texttt{\# Input}\\
\texttt{- `original\_code`: Raw LaTeX source.}\\
\texttt{- `regex\_reference`: Preliminary extraction by regex (prone to errors, not all of the results are trusted, must be carefully checked according to the original code).}

\texttt{\# Task Instructions}\\
\texttt{1. Analyze Structure: Identify if it's a TikZ figure (uses tikzpicture environment or inputs .tikz files). Identify if it's a multi-figure layout (multiple distinct images, subfigures, or minipages).}\\
\texttt{2. Resolve Macros (CRITICAL): Look for \textbackslash newcommand or \textbackslash def definitions within the code snippet. You MUST apply these substitutions to extracted paths.}\\
\texttt{3. Extract Paths: Extract clean file paths from \textbackslash includegraphics or \textbackslash input. CLEANING: Ignore LaTeX options like width=..., height=..., keepaspectratio. If a macro is external/unknown, keep the macro syntax.}\\
\texttt{4. Clean Captions: Extract the main caption text. Remove \textbackslash label\{\} and layout commands. Extract subcaptions if present.}

\texttt{\# Output JSON Schema}\\
\texttt{\{reasoning, is\_tikz, has\_multifigure, image\_paths, caption, subcaptions\}}
\end{tcolorbox}

\noindent\textbf{Primary VLM filter (Stage~2).}
\begin{tcolorbox}[title={Primary Filter --- System Prompt}, fonttitle={\bfseries\footnotesize}, fontupper=\scriptsize, colback={gray!5}, colframe={gray!60}, coltitle={white}]
\texttt{You are an expert vision-language classifier specialized in Academic Figure Analysis. Your task is to filter for high-quality, professional Method/Architecture Diagrams and strictly reject low-quality or irrelevant visuals.}

\texttt{TARGET CLASS: ``Methodologies, Architectures \& Conceptual Mechanics'' (ACCEPT)}\\
\texttt{The image serves as a visual explanation of the algorithm's inner workings.}\\
\texttt{* Professional Method/Architecture Diagrams: Clear, structured visualizations of system pipelines, neural networks, or algorithms.}\\
\texttt{* Explicit Connectivity: Must contain clear connectors (arrows, lines) showing data flow between distinct functional blocks.}\\
\texttt{* High Visual Quality: The image must be sharp and legible. You should be able to distinguish blocks, labels, and flow directions easily.}\\
\texttt{* Aesthetic Standard: Clean, vivid vector-style and professional diagrams. Logical spatial arrangement.}

\texttt{NEGATIVE (DROP - Reject if ANY apply):}\\
\texttt{* Low Quality/Corrupted: Blurry, pixelated, extreme crops, or ``black box'' images. Prioritize visual clarity over caption description.}\\
\texttt{* Misleading Captions: If the caption describes a method but the visual is just a corrupted crop or a simple icon, it is a REJECT.}\\
\texttt{* Experimental Results: Qualitative galleries, charts (X/Y axes), or comparison tables.}\\
\texttt{* Abstract/Messy 3D Scans: Chaotic 3D point clouds or raw scene reconstructions that lack structured ``flowchart'' elements.}\\
\texttt{* Text/Math-Major: Figures that are primarily pseudocode, long lists of text, prompt demonstrations or dense equations without aesthetic icons or structural boxes/arrows.}

\texttt{Output ONLY the JSON Schema:}\\
\texttt{\{``content\_analysis'': str, ``is\_target\_type'': bool\}}
\end{tcolorbox}

\noindent\textbf{Secondary quality inspector (Stage~2).}
\begin{tcolorbox}[title={Secondary Filter --- System Prompt}, fonttitle={\bfseries\footnotesize}, fontupper=\scriptsize, colback={gray!5}, colframe={gray!60}, coltitle={white}]
\texttt{You are a highly critical Scientific Figure Quality Inspector. Your mission is to identify professional-grade Method/Architecture Diagrams and ELIMINATE ``Dead/Broken Diagrams'' that failed to render or lack substantive content.}

\texttt{ACCEPTANCE CRITERIA (TARGET CLASS):}\\
\texttt{* Topological Complexity: Must have clear, visible arrows or directed edges connecting specific nodes.}\\
\texttt{* High Structural Density: Professional diagrams usually feature multiple stages, sub-modules, or complex data-flow paths (residual connections, skip links).}\\
\texttt{* Substantive Content: Each block/node MUST contain legible architecture details or text labels. If blocks are too abstractive, ``empty white boxes,'' it is a RENDER FAILURE.}\\
\texttt{* Professional Aesthetics: Clean vector graphics style with well-defined boundaries, with both icons and specific visual elements (not just plain boxes).}\\
\texttt{* Self-Explanatory: Even without the caption, the visual structure alone should suggest a sequence, hierarchy, or data flow.}

\texttt{REJECTION CRITERIA (REJECT if ANY apply):}\\
\texttt{* Render Failure (TikZ Artifacts): Images with empty rectangles or placeholders that don't convey the architecture. If the boxes are ``White Holes'' (missing text/math symbols inside), it's a render failure.}\\
\texttt{* Instructional/Prompt Demos: Figures with most area displaying text prompts, input-output examples, or UI screenshots. Figures that are primarily pseudocode or long lists of text.}\\
\texttt{* Non-Methodological Visuals: Qualitative galleries, charts, quantitative plots, histograms, 3D meshes/point clouds, screenshot/UI interfaces, biological/chemical cycles.}\\
\texttt{* Visual Corruption: Excessive cropping, severe blurring, or images where text and boxes are unidentifiably overlapped.}\\
\texttt{* Information Vacuity: Diagrams that are too simple (e.g., only 2-3 boxes with no internal logic) or ``skeleton'' versions that look unfinished.}

\texttt{Be extremely skeptical. If an image looks like a ``broken placeholder'' or a ``compilation error,'' mark is\_target\_type as false immediately.}
\end{tcolorbox}

\noindent\textbf{Topology-aware captioning (Stage~3).}
\begin{tcolorbox}[title={Captioning --- System Prompt}, fonttitle={\bfseries\footnotesize}, fontupper=\scriptsize, colback={gray!5}, colframe={gray!60}, coltitle={white}]
\texttt{TASK:}\\
\texttt{Given a research methodological figure, write a precise textual description (fig\_desc) so that the diagram can be redrawn without reading the paper.}

\texttt{Your description must contain the three aspects:}\\
\texttt{[1] Global Layout and Structure}\\
\texttt{[2] Visual Modules and Attributes}\\
\texttt{[3] Connections and Arrows}

\texttt{RULES:}\\
\texttt{1. Abstract the method the image depicts, and describe the overall layouts.}\\
\texttt{2. According to the methodology, introduce the workflow in its logical way, detailing the dominant models and their attributes (color, shape, binding text).}\\
\texttt{3. Pay attention to captions, subcaptions, and LaTeX equations to ensure knowledge completeness.}

\texttt{OUTPUT FORMAT:}\\
\texttt{a json containing a single key ``fig\_desc'' with the detailed description as value. Please use natural language, no markdown.}
\end{tcolorbox}

\begin{tcolorbox}[title={Captioning --- User Prompt}, fonttitle={\bfseries\footnotesize}, fontupper=\scriptsize, colback={gray!5}, colframe={gray!60}, coltitle={white}]
\texttt{You are provided with a high-fidelity academic architecture for a computer science paper.}\\
\texttt{Please generate fig\_desc NO MORE THAN 2000 TOKENS according to the system instructions.}

\texttt{ADDITIONAL CONTEXT:}\\
\texttt{- **LaTeX Context**: [LATEX\_FILLIN]}\\
\texttt{- **Figure Caption**: [CAPTION\_FILLIN]}
\end{tcolorbox}

\noindent\textbf{SAM3 mask verification (Stage~4).}
\begin{tcolorbox}[title={SAM3 Mask Verification --- System Prompt}, fonttitle={\bfseries\footnotesize}, fontupper=\scriptsize, colback={gray!5}, colframe={gray!60}, coltitle={white}]
\texttt{You are a helpful assistant specializing in detail-oriented visual understanding, reasoning, and classification, capable of carefully analyzing a predicted segmentation mask on an image along with zoomed-in views of the area around the predicted segmentation mask to determine whether the object covered by the predicted segmentation mask is one of the correct masks that match the user query.}

\texttt{The user will provide you with four pieces of information for you to jointly analyze before constructing your final prediction:}\\
\texttt{1. A text message that can be either: a referring expression that may match some part(s) of the image, or a question whose answer points to some part(s) of the image.}\\
\texttt{2. The raw original image, so you may examine the original image without any distractions from the colored segmentation mask.}\\
\texttt{3. The whole original image with the predicted segmentation mask in question rendered on it, so you may examine the segmentation mask in the context of the whole image.}\\
\texttt{4. A zoomed-in version of the predicted segmentation mask in question. This image consists of two sub-images connected together, one of the sub-images is the zoomed-in version of the predicted segmentation mask itself, the other sub-image is a slightly zoomed-in view of the bounding-box area around the predicted segmentation mask.}

\texttt{Now, please analyze the image and think about whether the predicted segmentation mask is a part of the correct masks that matches with or answers the user query or not. First output your detailed analysis of each input image, and then output your step-by-step reasoning, and then finally respond with either <verdict>Accept</verdict> or <verdict>Reject</verdict>.}
\end{tcolorbox}

\subsection{Evaluation Prompts}
\label{app:eval_prompt_templates}

\noindent\textbf{Reproducibility settings.} All VLM-based schema evaluation in this work uses deterministic decoding when available (temperature $=0$). We lock the evaluator prompt template, scoring rubric, and model snapshot identifier for each reported run. The same schema protocol is compatible with both deployment models and hosted large-model APIs; when an endpoint does not expose temperature control, we use its deterministic/lowest-variance decoding mode and record endpoint version and request configuration.

\noindent\textbf{Stage~1: Element Extraction Prompt.}
\begin{tcolorbox}[title={Stage~1: Element Extraction --- System + User}, fonttitle={\bfseries\footnotesize}, fontupper=\scriptsize, colback={gray!5}, colframe={gray!60}, coltitle={white}]
\textbf{System:}\\
\texttt{You are a scientific diagram structure extractor. Given a reference diagram and its caption, identify all structural elements. Be exhaustive but precise --- list every distinct component, arrow/connection, and text label visible in the image.}

\textbf{User:}\\
\texttt{Analyze this scientific diagram and extract ALL structural elements.}\\
\texttt{Caption: ``\{caption\}''}

\texttt{List every element in these categories:}\\
\texttt{1. **Components**: Named blocks, shapes, modules, icons (anything that is a distinct visual unit)}\\
\texttt{2. **Arrows / Connections**: Directed or undirected connections between components}\\
\texttt{3. **Text Labels**: All text visible in the diagram (labels, annotations, equations)}

\texttt{Respond ONLY with a JSON object:}\\
\texttt{\{``components'': [\{``name'': ..., ``description'': ...\}],}\\
\texttt{~``arrows'': [\{``source'': ..., ``target'': ..., ``label'': ...\}],}\\
\texttt{~``text\_labels'': [\{``text'': ..., ``location'': ...\}]\}}
\end{tcolorbox}

\noindent\textbf{Stage~2: Dimension-Specific Error Detection.}
For SciFormaBench-2K, evaluation is decomposed into three independent API calls (one per dimension) to prevent context exhaustion and ensure focused verification. Below we provide the exact system and user prompts for Components, Arrows, and Text.

\begin{tcolorbox}[title={Stage~2: Component Dimension --- System + User}, fonttitle={\bfseries\footnotesize}, fontupper=\scriptsize, colback={gray!5}, colframe={gray!60}, coltitle={white}, breakable]
\textbf{System:}\\
\texttt{You are a scientific diagram quality evaluator. You will see TWO images and a list of components extracted from the reference. LEFT = REFERENCE (ground truth). RIGHT = GENERATED (to evaluate). Your ONLY job is to check components/blocks/shapes. Do NOT report minor stylistic differences or speculative issues.}\\

\textbf{User:}\\
\texttt{\#\# Context}\\
\texttt{You are evaluating a GENERATED scientific diagram. Focus ONLY on components.}\\
\texttt{The diagram should depict: ``\{prompt\}''}\\
\texttt{LEFT image = REFERENCE. RIGHT image = GENERATED.}\\

\texttt{\#\# Reference Components (\{n\_comp\} total)}\\
\texttt{\{comp\_list\}}\\

\texttt{\#\# Your Task}\\
\texttt{Two checks:}\\
\texttt{1. **Reference check**: For each component listed above, check if it is absent/wrong.}\\
\texttt{2. **Hallucinated check**: Identify any major components that are clearly wrong or noise.}\\

\texttt{Error types (in priority order — use the FIRST matching type):}\\
\texttt{1. ``Missing'': a reference component is entirely absent $\to$ critical}\\
\texttt{2. ``Hallucinated'': a major element that clearly should not exist $\to$ critical}\\
\texttt{3. ``Distorted'': shape is severely distorted, blurred, or illegible $\to$ critical}\\
\texttt{4. ``Structural mismatch'': internal structure differs significantly from the reference $\to$ moderate}\\
\texttt{5. ``Duplicate'': appears more times than in reference $\to$ moderate}\\
\texttt{6. ``Wrong'': present but depicts a clearly different concept $\to$ moderate}\\

\texttt{Report ONLY the types listed above. Start each error description with the type name.}
\end{tcolorbox}

\begin{tcolorbox}[title={Stage~2: Arrow Dimension --- System + User}, fonttitle={\bfseries\footnotesize}, fontupper=\scriptsize, colback={gray!5}, colframe={gray!60}, coltitle={white}, breakable]
\textbf{System:}\\
\texttt{You are a scientific diagram quality evaluator. You will see TWO images and a list of arrows/connections extracted from the reference. LEFT = REFERENCE. RIGHT = GENERATED. Your job is to check arrows/connections for missing and hallucinated errors. Do NOT report minor stylistic differences.}\\

\textbf{User:}\\
\texttt{\#\# Context}\\
\texttt{You are evaluating a GENERATED diagram. Focus ONLY on arrows/connections.}\\
\texttt{The diagram should depict: ``\{prompt\}''}\\

\texttt{\#\# Reference Arrows / Connections (\{n\_arrow\} total)}\\
\texttt{\{arrow\_list\}}\\

\texttt{\#\# Your Task}\\
\texttt{Two checks: 1. **Missing check**. 2. **Hallucinated check**.}\\

\texttt{Error types:}\\
\texttt{- ``Missing'': a reference connection is entirely absent $\to$ critical}\\
\texttt{- ``Hallucinated'': an arrow that clearly should not exist $\to$ critical}\\
\texttt{- ``Position error'': start or end point is noticeably off $\to$ critical}\\
\texttt{- ``Wrong'': connects clearly wrong components or wrong direction $\to$ moderate}\\

\texttt{Do NOT report routing differences, curvature, thickness, or color errors.}\\
\texttt{Start the description with the error type.}
\end{tcolorbox}

\begin{tcolorbox}[title={Stage~2: Text Dimension --- System + User}, fonttitle={\bfseries\footnotesize}, fontupper=\scriptsize, colback={gray!5}, colframe={gray!60}, coltitle={white}, breakable]
\textbf{System:}\\
\texttt{You are a scientific diagram quality evaluator specialising in text readability. LEFT = REFERENCE. RIGHT = GENERATED. Diffusion models frequently produce garbled text. Catch every such case.}\\

\textbf{User:}\\
\texttt{\#\# Context}\\
\texttt{Focus ONLY on text labels. Diffusion models frequently struggle with text rendering. Pay special attention to whether each label is actually legible.}\\
\texttt{The diagram should depict: ``\{prompt\}''}\\

\texttt{\#\# Reference Text Labels (\{n\_text\} total)}\\
\texttt{\{text\_list\}}\\

\texttt{\#\# Your Task}\\
\texttt{For every text label, check: (1) is it present? (2) is it legible? (3) is it correct?}\\

\texttt{Error types:}\\
\texttt{1. ``Missing'': label is entirely absent $\to$ critical}\\
\texttt{2. ``Garbled / unreadable'': character-level corruption (misspelled, distorted) $\to$ critical}\\
\texttt{3. ``Truncated'': label is cut off or incomplete $\to$ critical}\\
\texttt{4. ``Wrong text'': fully legible but says something completely different $\to$ moderate}\\
\texttt{5. ``Duplicated'': same label appears where it shouldn't $\to$ moderate}\\

\texttt{If a label has ANY character corruption, classify as Garbled. Do not penalise text added by the generator.}
\end{tcolorbox}

\noindent\textbf{M-DPO Reward Scoring Prompt.}
During preference-pair construction (Section~\ref{app:preference_construction}), we additionally use a compact VLM scoring prompt for rapid candidate ranking:

\begin{tcolorbox}[title={Reward Scoring --- System Prompt}, fonttitle={\bfseries\footnotesize}, fontupper=\scriptsize, colback={gray!5}, colframe={gray!60}, coltitle={white}]
\texttt{You are a technical diagram analyst. Compare generated diagram (RIGHT) against reference (LEFT) for SEMANTIC errors only.}

\texttt{WHAT IS AN ERROR (report these):}\\
\texttt{- Text: Misspellings, wrong terms, missing/garbled labels}\\
\texttt{- Arrows: Wrong logical connection (A->B should be A->C), missing critical flow, extra spurious connection}\\
\texttt{- Components: Missing module, extra module, wrong logical order in data flow}

\texttt{WHAT IS NOT AN ERROR (DO NOT report):}\\
\texttt{- Layout differences (vertical vs horizontal, left vs right placement)}\\
\texttt{- Spacing, alignment, visual styling differences}\\
\texttt{- Color or shape variations}\\
\texttt{- Position changes that preserve the same logical flow}

\texttt{OUTPUT (strict JSON):}\\
\texttt{\{``reference\_description'': \{``overview'': ..., ``components'': [...], ``arrows'': [...]\},}\\
\texttt{~``generated\_description'': \{...\},}\\
\texttt{~``errors'': \{``text\_errors'': [...], ``arrow\_errors'': [...], ``component\_errors'': [...]\},}\\
\texttt{~``assessment'': \{``quality\_score'': 0--10, ``summary'': ...\}\}}
\end{tcolorbox}

\begin{tcolorbox}[title={Reward Scoring --- Refinement Prompt}, fonttitle={\bfseries\footnotesize}, fontupper=\scriptsize, colback={gray!5}, colframe={gray!60}, coltitle={white}]
\texttt{Review your previous analysis and REMOVE any errors that are just layout/position differences:}

\texttt{Previous analysis: \{previous\_analysis\}}

\texttt{REMOVE these types of non-errors:}\\
\texttt{- ``placed under instead of left/right'' - layout difference, not error}\\
\texttt{- ``vertical vs horizontal'' - layout difference}\\
\texttt{- ``spatial flow'' changes - if logical flow A->B->C is preserved, not an error}\\
\texttt{- ``wrong\_order'' where the logical sequence is actually preserved}

\texttt{KEEP only TRUE semantic errors:}\\
\texttt{- Text misspellings, wrong terms}\\
\texttt{- Missing/extra logical connections}\\
\texttt{- Missing/extra components}

\texttt{Output the CORRECTED JSON with only true errors remaining.}
\end{tcolorbox}

\bibliographystyle{ACM-Reference-Format}
\bibliography{sciforma}

\end{document}